\documentclass[lettersize,journal]{IEEEtran}
\usepackage{amsmath,amsfonts}
\usepackage{algorithmic}
\usepackage{algorithm}
\usepackage{array}
\usepackage[caption=false,font=normalsize,labelfont=sf,textfont=sf]{subfig}
\usepackage{textcomp}
\usepackage{stfloats}
\usepackage{url}
\usepackage{verbatim}
\usepackage{graphicx}
\usepackage{adjustbox}
\usepackage{cite}
\usepackage{booktabs}
\usepackage[table,xcdraw]{xcolor}
\begin{document}

\title{Image Harmonization with Region-wise \\ Contrastive Learning}

\author{Jingtang~Liang
and Chi-Man~Pun,~\IEEEmembership{Senior~Member,~IEEE}
\thanks{Corresponding author: Chi-Man~Pun}

}

\markboth{Journal of \LaTeX\ Class Files,~Vol.~14, No.~8, August~2021}%
{Shell \MakeLowercase{\textit{et al.}}: A Sample Article Using IEEEtran.cls for IEEE Journals}

\IEEEpubidadjcol

\maketitle

\begin{abstract}
Image harmonization task aims at harmonizing different composite foreground regions according to specific background image. Previous methods would rather focus on improving the reconstruction ability of the generator by some internal enhancements such as attention, adaptive normalization and light adjustment, $etc.$. However, they pay less attention to discriminating the foreground and background appearance features within a restricted generator, which becomes a new challenge in image harmonization task.
In this paper, we propose a novel image harmonization framework with external style fusion and region-wise contrastive learning scheme. For the external style fusion, we leverage the external background appearance from the encoder as the style reference to generate harmonized foreground in the decoder. This approach enhances the harmonization ability of the decoder by external background guidance. Moreover, for the contrastive learning scheme, we design a region-wise contrastive loss function for image harmonization task. Specifically, we first introduce a straight-forward samples generation method that selects negative samples from the output harmonized foreground region and selects positive samples from the ground-truth background region. Our method attempts to bring together corresponding positive and negative samples by maximizing the mutual information between the foreground and background styles, which desirably makes our harmonization network more robust to discriminate the foreground and background style features when harmonizing composite images. 
Extensive experiments on the benchmark datasets show that our method can achieve a clear improvement in harmonization quality and demonstrate the good generalization capability in real-scenario applications. 
\end{abstract}
\section{Introduction}
\IEEEPARstart{I}{mage} composition aims at combining different regions (denoted as foreground) from other images into a different background, composing a new composite image. However, here comes a more straightforward problem that the composite image inevitably suffers from the disharmonious appearance between the foreground and background regions caused by color, illuminations, $etc.$, which can be visually distinguished by the human eyes. To address the inconsistent appearance, image harmonization task specializes in making foreground regions more compatible and realistic with background regions, which also has potential to solve contemporary problems in image editing~\cite{editing1, encoder3}, relighting~\cite{relighting}, enhancement\cite{enhancement}, synthesis\cite{synthesis1, synthesis3, synthesis2}, forgery detection~\cite{foregy1,foregy2,forgery3}  and video synthesis\cite{video_syn, lee2019inserting}, $etc.$.

The key solution addressing image harmonization task is to adjust foreground appearance to make it more compatible with the corresponding background while also maintaining the unchanged structure and semantic information. Traditional harmonization methods intend to transfer handcrafted statistics and low-level features, such color~\cite{intro_color1,intro_color2}, illumination~\cite{intro_illumination} and texture~\cite{intro_texture}, $etc.$. While traditional solutions achieve unreliable performance in vastly different scenarios, several deep-learning based harmonization models have been proposed to address this challenging task. A number of recent works mainly concentrate on exploring the associations between the foreground and background regions, such as designing attention module to learn regional features~\cite{s2am}, employing domain discriminator to identify the composite and natural foreground regions~\cite{dovenet}, extracting domain-aware guidance for harmonization~\cite{bargainet} and adaptively transferring background style into foreground regions~\cite{Ling_2021_CVPR}. Above methods achieve promising results in image harmonization and particularly benefit from the strong reconstruction ability of the employed encoder-decoder UNet~\cite{unet} architecture. 

However, after extracting the context of composite image by the encoder in UNet, the frameworks in previous approaches not only confronts limited ability to identify the foreground and background features, but also shows weakness in harmonizing the foreground region with background style only under supervision of the ground-truth image. In this paper, we propose an alternative, rather straightforward approach to address the above problems. Given specific patches in the foreground region, the style of the patches should be naturally consistent with the background patches for harmonization. To do so, we need to learn and focus on the appearance consistency between the two domains, the foreground and background, while being invariant to the necessarily different features including textures and structures in various regions. Hence, we achieve this by employing a kind of contrastive loss function, InfoNCE loss\cite{infonce}, which we redesign to associate the style of the foreground and background patches to each other for the first time. In this work, we firstly use it at a region-wise supervised scheme, where we simultaneously conduct multiple negative samples from the harmonized foreground and multiple positive samples from the ground-truth background. We find that drawing negative and positive samples internally and regionally from the respective foreground and background regions, rather than generally selecting samples from the whole image, enforces the decoder to generate foregrounds with more harmonious improvement. 

Furthermore,  we find that most of the existing methods attempt to improve the reconstruction ability  by enhancing internal information (attention module~\cite{s2am},  domain information~\cite{bargainet} and normalization~\cite{ Ling_2021_CVPR}) only within the decoder to avoid style information decreases.  Differently, as several researches~\cite{encoder1,encoder2,encoder3} show that the encoder will become stabilized when extracting data-driven features during the training process,  we approach to add appearance guidance from the encoder. Compared to the unstable background features that are reconstructed from the decoder, the unchanged background features convey more accurate region-specific details when extracting from the encoder and thus contribute to greater harmonization. The background appearance guidance~(from encoder) and the reconstructed foreground appearance~(from decoder) are further embedded and fused in the decoder via style fusion. This straightforward scheme shows a clear improvement in harmonization quality, without large computational cost. Extensive experiments demonstrate that the our approaches introduced above can effectively conspire together not only in  discriminating the foreground and background features in a novel scheme, but also in further improving the harmonization quality and outperforming the state-of-the-art methods both on benchmark synthesis and real-scenario datasets. 

Overall, the main contributions of this paper can be summarized as follows:

\begin{itemize}
    \item We propose a novel \textbf{region-wise} contrastive loss function for  discriminating the foreground and background features and further facilitating the harmonization model to learn better regional features.
    \item We \textbf{first} employ background style representations in feature extractor to guide the harmonization generator via the proposed external style fusion strategy, further improving harmonization quality of the foreground region.
    \item Experimental results show that our method outperforms other state-of-the-art methods qualitatively and quantitatively on the benchmark datasets and also achieves favorable results in real-scenario applications.
\end{itemize}

\section{Related Work}
\textbf{Image Harmonization}
Traditional image harmonization methods mainly focus on adjusting low-level features and handcrafted appearance of the foreground and background regions, such as adjusting global color distributions \cite{rw_color1,rw_color2}, applying gradient domain composition \cite{rw_gradient1,rw_gradient2} or manipulating multi-scale transformation and statistical analysis \cite{intro_texture}. Although above methods achieve preliminary results in image harmonization task, the quality of the coarse harmonized image cannot be visually guaranteed. Recently, several deep learning methods have been further investigated in solving image harmonization task. For instance, \cite{zhu2015learning} first propose a pre-trained discriminator model to adjust the appearance of the composite images.  Later, a few researches claim that the key solution to harmonize composite image is to maintain semantic consistency~\cite{DIH_Tsai, sofiiuk2021foreground}. Current methods in image harmonization task gradually focus on figuring out the differences between the foreground and background region features, such as employing attention module to learn regional spatial features~\cite{s2am}, designing domain verification discriminator to identify the foreground and background regions~\cite{dovenet}, adding domain-aware guidance\cite{bargainet}, redefining mask-guided spatial normalization~\cite{Ling_2021_CVPR, Guo_2021_CVPR} and leveraging transformer to obtain regional consistency\cite{ihtrans}. However, while above methods seem to enhance internal information only within their employed UNet decoder, their models reveal limitations not only in effectively identify the styles between foreground and background regions but also in reconstructing appropriate foreground appearance according to corresponding background appearance.

\textbf{Contrastive learning.} Contrastive learning has been proven as an effective method in unsupervised visual representation learning. Tracing back to~\cite{contrastive1}, their methods first learn representations by contrasting positive pair against negative pairs. This scheme drives~\cite{infonce} to explore the latent space and propose a probabilistic contrastive loss to capture maximal mutual information between the positive and negative pairs, which is also an effective approach adopted and extended in several papers~\cite{moco, byol, simclr, liu2022contrastive}.

As the contrastive learning greatly benefits the visual representation learning, some researches have explored its efficiency in many downstream tasks. For example, CUT~\cite{patchnce} is the pioneer contrastive learning approach in image-to-image translation tasks. Their approaches are to propose a patch-based contrastive loss to match style consistency between the positive samples and negative samples. Later, DCL-GAN~\cite{dclgan} introduced a bidirectional patch-based contrastive loss to match the source and target domain. Additionally in super-resolution task, \cite{contrastive4} and \cite{contrastive5} have successfully attempted to embed the low-resolution and high-resolution samples into latent representations to obtain maximal mutual information. The designed choices of defining positive and negative samples play an essential and critical role in various image translation tasks. 
In this paper, we approach to investigate contrastive learning  for image harmonization task by proposing a novel region-wise contrastive loss function.

\section{Proposed Method}
\label{sec:method}
In this section, we first introduce the base network design to generate harmonized image. Then we describe how the proposed external style fusion block to provide background appearance guidance. After that, we mainly focus on illustrating our proposed region-wise contrastive learning scheme for image harmonization and explaining our strategy to generate positive and negative samples from foreground and background regions respectively. At last we introduce our loss functions.  

\subsection{Image Harmonization Network with External Style Fusion}
\begin{figure*}[t]
    \centering
    \includegraphics[width=\textwidth]{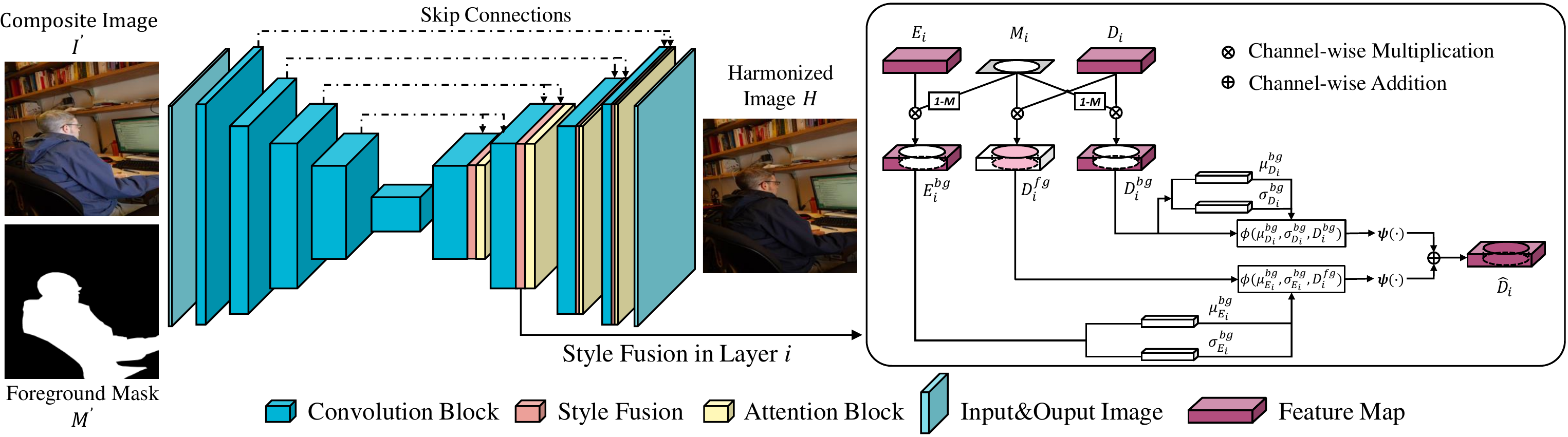}
    \caption{The overview of our image harmonization network with external style fusion. The base network takes low-resolution composite image  $I^{'}$ and foreground mask $M^{'}$ as inputs and generates the harmonized image $H$.
    The style fusion in layer \emph{i} employs the background appearance from the encoder as style reference, conspires with AdaIN~\cite{adain} and follows the flow chart to obtain the augmented feature map $\widehat{D_{i}}$ in layer \emph{i}. Notice that the shapes of $E_{i}$ and $D_{i}$ are symmetrically equal. }
    \label{fig:nn}
\end{figure*}

\textbf{Base Network.}
Formally, given the original composite image $ I \in \mathbb{R}^{3 \times H \times W} $ and its corresponding foreground mask $ M \in \mathbb{R}^{1 \times H \times W} $, we first downsample $I$ and $M$  to obtain low-resolution $I^{'}$ and $M^{'}$. Our base network takes $I^{'}$ and $M^{'}$ as inputs and generates harmonized image $ H \in \mathbb{R}^{3 \times H \times W} $, which consists of four modules: encoder $G_{enc}$, decoder $G_{dec}$, attention module and style fusion layer. We employ a encoder-decoder based residual-UNet~(ResUNet)~\cite{resUnet} as $G_{enc}$ and $G_{dec}$, which has been proven to better expand the receptive field and improve the quality of the recovered image. For attention module, we employ S$^2$AM~\cite{s2am} following previous approaches~\cite{bargainet, Ling_2021_CVPR}. Our decoder network $G_{dec}$ is inserted with S$^2$AM module to enhance internal spatial information and style fusion layer to provide external style guidance. The whole architecture is illustrated in Figure~\ref{fig:nn}. 

\textbf{External Style Fusion.}
With $G_{enc}$ and $G_{dec}$, we can obtain encoder feature maps $\{E_{i}\}=G_{e n c}\left(I^{'}\right)$ and decoder feature maps $\{D_{i}\}=G_{d e c}\left(G_{e n c}\left(I^{'}\right)\right)$, where \emph{i} denotes the \emph{i}-th layer. As the encoder and decoder are structurally symmetric, we reverse the order of encoder feature maps $\{E_{i}\}$ to equate the shapes of $D_i$ and $E_{i}$. Preliminarily, we divide the foreground and background regions in each feature map by multiplying the re-scaled foreground mask $M_i$ and background mask $(1-M_i)$ in order to learn the foreground and background styles separately, which are formulated by:
\begin{equation}
E_i^{b g}=G_{e n c}\left(I^{\prime}\right) \times\left(1-M_i\right)
\end{equation}
\begin{equation}
D_i^{b g}=G_{dec}\left(G_{enc}\left(I^{\prime}\right)\right) \times\left(1-M_i\right)
\end{equation}
\begin{equation}
D_i^{f g}=G_{dec }\left(G_{e n c}\left(I^{\prime}\right)\right) \times M_i
\end{equation}
where $E_i^{b g}$ and $D_i^{b g}$ are \emph{i}-th background feature map in $G_{enc}$ and $G_{dec}$, and $D_i^{f g}$ is \emph{i}-th foreground feature map in $G_{dec}$. Unlike the internal normalization in RAIN~\cite{Ling_2021_CVPR}, we particularly leverage the external style guidance from the background of $E_{i}$ instead of $D_{i}$. To this end, we calculate the channel-wise mean $\mu_{E_i}^{b g}$ and standard variance $\sigma_{E_i}^{b g}$ of the background feature map $E_i^{b g}$. Particularly, we normalize the foreground region in $D_i^{f g}$ to preliminarily harmonize the composite foreground regions, where the the normalization in layer-\emph{i} is formulated by:
\begin{equation}
\widehat{D_i^{f g}}=\phi\left(D_i^{f g}, \mu_{E_i}^{b g}, \sigma_{E_i}^{b g}\right)=\frac{D_i^{f g}-\mu_{E_i}^{b g}}{\sigma_{E_i}^{b g}}
\end{equation}
To the best of our knowledge, we are the first approach to leverage the style features in encoder to guide the reconstruction of harmonious foreground, while previous methods use the basic normalization\cite{bargainet, Guo_2021_CVPR} or style transfer~\cite{Ling_2021_CVPR} only within the decoder. Similarly, we calculate the channel-wise mean $\mu_{D_i}^{b g}$ and standard variance $\sigma_{D_i}^{b g}$ of the background feature map $D_i^{b g}$, and next normalize the the background region using the analogical formula:
\begin{equation}
\widehat{D_i^{b g}}=\phi\left(D_i^{b g}, \mu_{D_i}^{b g}, \sigma_{D_i}^{b g}\right)=\frac{D_i^{b g}-\mu_{D_i}^{b g}}{\sigma_{D_i}^{b g}}
\end{equation}

Notice that for background normalization, we leverage the internal background style guidance only from $D_i^{b g}$ for the trade-off between reconstruction and over-fitting. After we obtain the foreground feature map $\widehat{D_{i}^{f g}}$ and background feature map $\widehat{D_{i}^{b g}}$, we adopt a weighted fusion operation $\psi$ to filter out inaccurate appearance details, which facilitates the $\widehat{D_{i}^{f g}}$ and $\widehat{D_{i}^{b g}}$  to produce the augmented foreground and background maps by:
\begin{equation}
\psi(\widehat{D_{i}^{fg}})=f_{i}^{s}(\widehat{D_{i}^{fg}}) \times \widehat{D_{i}^{fg}}+f_{i}^{b}(\widehat{D_{i}^{fg}})
\end{equation}
\begin{equation}
\psi(\widehat{D_{i}^{bg}})=f_{i}^{s}(\widehat{D_{i}^{bg}}) \times \widehat{D_{i}^{bg}}+f_{i}^{b}(\widehat{D_{i}^{bg}}),
\end{equation}
where $f_{i}^{s}$ and $f_{i}^{b}$ are affine functions for calculating the scale $s_i$ and bias $b_i$ in AdaIN~\cite{adain}.  Finally we obtain the \emph{i}-th augmented feature map $\widehat{D_{i}}$ in decoder $G_{dec}$ by channel addition:
$
\widehat{D_{i}}=\psi(\widehat{D_{i}^{fg}})\oplus \psi(\widehat{D_{i}^{bg}}).
$
The entire style fusion procedure is briefly introduced in Figure~\ref{fig:nn}. All the hyper-parameters are initialized as 0 and automatically optimized during the training process.

\subsection{Region-wise Contrastive Learning Approach}
With the augmented feature maps $\{\widehat{D_{i}}\}$, we sequentially apply $\{\widehat{D_{i}}\}$ into each layer of the decoder and finally generate harmonized image \emph{H}. Existing approaches are mainly used to supervising \emph{H} with the ground-truth natural image $\widehat{H}$ by  pre-specific loss functions like $\mathcal{L}_{1}=\left\|H- \widehat{H} \right\|_{1}$ and $\mathcal{L}_{2}=\left\|H-\widehat{H}\right\|_{2}$. Although these conventional loss functions have achieved promising results in reconstructing desired images, but they also have limitations in some specific tasks that needs regional feedback like image harmonization. 

In this paper, we attempt to explore a more effective way to discriminate the difference between foreground and background appearance. Inspired by CUT~\cite{patchnce}, we first associate the foreground and background appearance as the negative and positive samples in contrastive learning. Different from CUT that conducts samples from the whole image, our method is more task-specific in generating positive and negative samples from different regions within the same image. In the following subsections we introduce the details of our contrastive learning approaches.

\begin{figure*}[t]
    \centering
    \includegraphics[width=\textwidth]{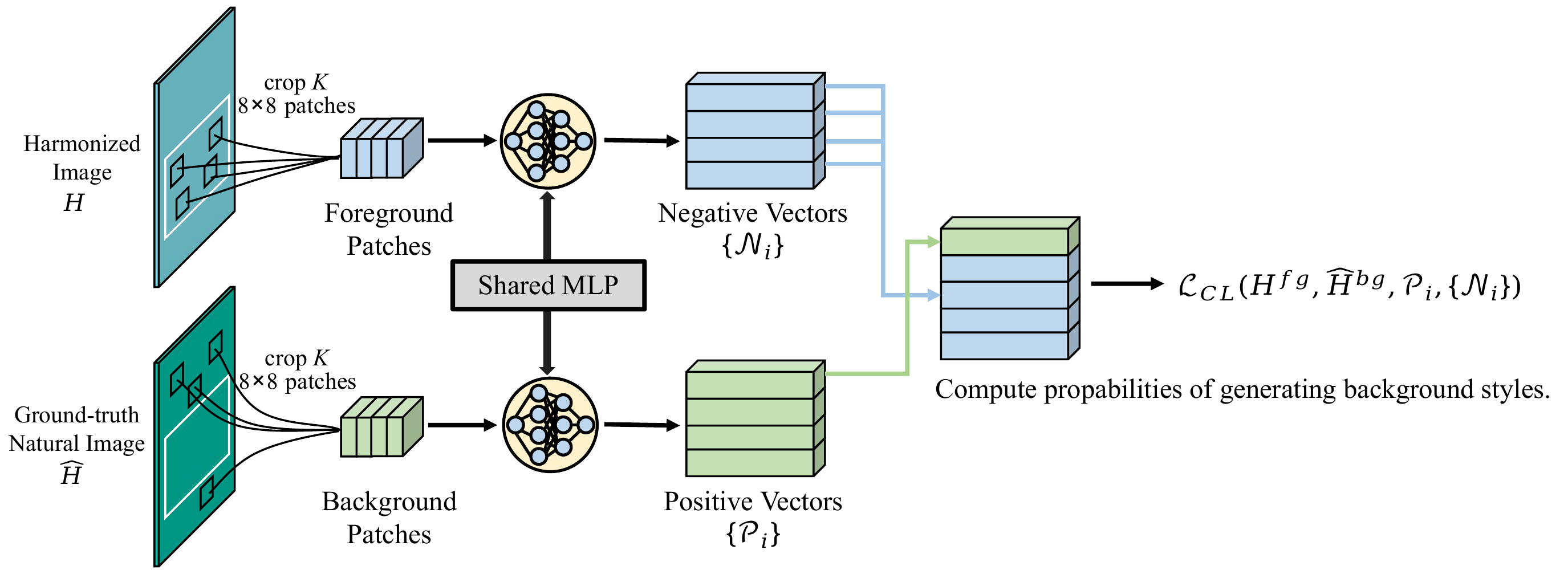}
    \caption{Negative patches from the foreground region of harmonized output $H$ and positive samples from the background region of ground-truth $\widehat{H}$ are embedded through a shared \emph{MLP} and compute the contrastive loss $\mathcal{L}_{CL}$. The while boxes in $H$ and $\widehat{H}$ represent the foreground regions.}
    \label{fig:contrastive}
\end{figure*}
\subsubsection{Region-wise Positive and Negative Samples Generation}
\label{sec:patches}
Contrastive learning has been proven to be effective both at image and patch level~\cite{contrastive3, contrastive6,dclgan}. Specifically in our harmonization task, we notice that not only both the complete foreground and background regions share similar style information~\cite{Ling_2021_CVPR}, but also the corresponding patches respectively from the foreground and background regions do. Since contrasting pixel-to-pixel information is impractical in comparing style information, we denote the foreground of harmonized image \emph{H} as negative and the background of ground-truth image $\widehat{H}$ as positive, and we take valid and region-specific strategies to generate our negative and positive patches.

After obtaining the harmonized image $ H \in \mathbb{R}^{3 \times H \times W} $, we perform matrix multiplication with its foreground mask $M^{'}$ to generate foreground region $H^{fg} \in \mathbb{R}^{3 \times H \times W} $. We first downsample $H^{fg}$ by factors $H/4$ and $W/4$ to obtain foreground sampling map $S^{fg}$, such that each pixel of sampling map correlates to a $4\times4$ region on original harmonized image $H$. Next, we randomly select \emph{K} pixels from $S^{fg}$ as negative locations and then crop $8\times8$ patches around the negative locations in $S^{fg}$. Hence, we embed the \emph{K} patches through a two-layer MLP following~\cite{patchnce} and denote that the embedding features compose the negative set $\{\mathcal{N}_i \} $, where each negative vector is defined as follows:
\begin{equation}
\mathcal{N}_i=\left\{S_i^{f g} \mid \mathcal{N}_i=MLP\left(\varphi\left(H^{f g} ; S_i^{f g}\right)\right)\right\}
\end{equation}
where $\varphi$ is the random cropping function and $S^{fg}_{i}$ denotes the \emph{i}-th negative locations sampling map. Similarly, with the background mask $(1-M^{'})$, we can obtain background region $\widehat{H}^{bg}$ from ground-truth image $\widehat{H}$. Following the above negative patches generation strategy, we also embed the patches through the same MLP to compose the positive set $\{\mathcal{P}_i \}$, where each positive vector is defined as follows:
\begin{equation}
\mathcal{P}_{i}=\left\{S_{i}^{ {bg }} \mid \mathcal{P}_{i}=MLP\left(\varphi\left(\widehat{H}^{bg} ; S_{i}^{ {bg }}\right)\right)\right\}
\end{equation}
One may have following questions: (1) \emph{Why not directly adopt the $H/8$ and $W/8$ down-sampling factor instead of $H/4$ and $W/4$?} As we empirically set the image size of the output harmonized image \emph{H} as $256 \times 256$, the size of sampling map will become $32\times32$ using factor $H/8$ and $W/8$, such that not only the sampling locations will decrease by a large margin~(from $64\times64$ to $32\times32$), but the overlaps of the cropped $8\times8$ patches also adjacently increase. (2)~\emph{The differences compared with CUT~(\cite{patchnce})?} As the CUT method selects patches from the whole image, it has limitation on selecting patches regionally and simultaneously according to different foreground masks in harmonization task while it restricts the patch locations both on positive and negative feature maps. 

\subsubsection{Contrastive Loss}
Given the background query feature map \emph{$v^{bg}$} with its positive set $\{\mathcal{P}_i\}$ and the foreground query feature map as \emph{$v^{fg}$} with its negative set $\{\mathcal{N}_i\}$, we compute the probability that a positive vector $\mathcal{P}_i$ will be selected over all the negative vectors $\{\mathcal{N}_i\}$, which practically means that the harmonization model would rather choose to generate background styles than foreground styles. Mathematically, above problem can be formulate as a cross-entropy loss:
\begin{equation}
\begin{aligned}
&\mathcal{L}_{C L}\left(v^{fg}, v^{bg}, \mathcal{P}_{i}, \{\mathcal{N}_i\} \right)= \\ 
&-\log \frac{\exp \left(v^{bg} \cdot \mathcal{P}_{i} / \tau\right)}{\exp \left(v^{bg} \cdot \mathcal{P}_{i} / \tau\right)+\sum_{n=1}^{K} \exp \left(v^{fg} \cdot \mathcal{N}_{n} / \tau\right)},
\end{aligned}
\end{equation}
where $\tau$ denotes a temperature parameter~\cite{infonce} and we use 0.07 as default. As we select \emph{K} patches in Section~\ref{sec:patches}, our contrastive loss for each pair of \emph{$v^{bg}$} and \emph{$v^{fg}$} is defined as follows:
\begin{equation}
\begin{aligned}
&\mathcal{L}_{C L}\left(v^{ {bg }}, v^{ {fg }}\right)= \\
&\frac{1}{K} \sum_{i=1}^{K}-\log \frac{\exp \left(v^{ {bg }} \cdot \mathcal{P}_{i} / \tau\right)}{\exp \left(v^{ {bg }} \cdot \mathcal{P}_{i} / \tau\right)+\sum_{k=1}^{K} \exp \left(v^{ {fg }} \cdot \mathcal{N}_{k} / \tau\right)}.
\end{aligned}
\end{equation}
Therefore, our total contrastive loss for training harmonization network is formulated as follows:
\begin{equation}
\mathcal{L}_{H C L}=\frac{1}{N} \sum_{i=1}^{N} \mathcal{L}\left(H_{i}^{ {fg }}, \widehat{H}_{i}^{ {bg }}\right),
\label{eq:clloss}
\end{equation}
where \emph{N} is the total number of the training images.

\subsection{Overall Loss Function}
We consider our method as a supervised problem. To achieve faithful harmonization, we calculate the $\mathcal{L}_{1}$ loss betwen the whole harmonized image~\emph{H} and the ground-truth image  $\widehat{H}$. We also calculate the $\mathcal{L}_{1}$ loss only within the foreground region~(denoted as $\mathcal{L}_{pixel}$) following~\cite{cun2020split} to learn better foreground appearance. At last, the above two loss functions and our proposed contrastive loss~(Eq.\ref{eq:clloss}) conspire together to compose the overall loss of our method, which is a weighted formulation defined as:
\begin{equation}
\mathcal{L}_{ {all }}=\lambda_{1} \mathcal{L}_{1}+\lambda_{2} \mathcal{L}_{ {pixel }}+\lambda_{3} \mathcal{L}_{H C L},
\label{eq:loss}
\end{equation}
where we empirically set $\lambda_{1}=0.4$, $\lambda_{2}=0.5$ and $\lambda_{3}=0.1$ for all our training experiments.


\section{Experiments}

\subsection{Implementation Details}
\label{exp:Imp_details}
\textbf{Settings.}
We implement our method in Pytorch~\cite{pytorch} and train on a single NVIDIA TITAN V GPU with 12GB memory. All the images are resized to 256$\times$256, random cropped and flipped for fair evaluation following previous methods~\cite{dovenet,bargainet,Ling_2021_CVPR}.  For training the whole network, we use batch size 8 and learning rate 1e-3 for convergence. We leverage the AdamW optimizer~\cite{adamw} with default setting and $K=256$ during all the training, where \emph{K} is the number of patches as introduced in Section~\ref{sec:method}. 

\subsection{Datasets}
\textbf{iHarmony4 synthesis dataset.}
The iHarmony4 dataset~\cite{dovenet} is a synthetic dataset that contains 73146 image pairs for image harmonization, including the synthetic images, the corresponding foreground masks and the target images. The iHarmony4 totally includes 65742 training images pairs and 7404 testing image pairs, and it consists of 4 sub-datasets~(HCOCO, HAdobe5k, HFlickr, Hday2night), the details of which are presented in Table~\ref{table:subset}. Following previous methods~\cite{dovenet,s2am,Ling_2021_CVPR,bargainet,Guo_2021_CVPR}, we primarily train our framework on the training samples of iHarmony4 and we preliminarily leverage the testing samples in iHarmony4 dataset to estimate the harmonization performance of our model on synthetic dataset.

\begin{table}[tbh!]
\centering
\caption{The details of 4 sub-datasets in iHarmony4.}
\begin{tabular}{@{}c|ccccc@{}}
\toprule
Sub-dataset & HCOCO & HAdobe5k & Hflickr & Hdat2night & Total \\ \midrule
Training    & 38545 & 19437    & 7449    & 311        & 65742 \\
Testing     & 4283  & 2160     & 828     & 133        & 7404  \\ \bottomrule
\end{tabular}
\label{table:subset}
\end{table}

\textbf{DIH real-scenario dataset.}
DIH real-scenario dataset\cite{DIH_Tsai}~(DIH99) is a real-scenario dataset that covers a variety of real-scenario composite examples. It contains 99 copy-paste high-quality composite images with the corresponding foreground masks that are collected from various real scenarios and stylized images. Following previous methods~\cite{dovenet,s2am,Ling_2021_CVPR,bargainet,Guo_2021_CVPR}, we conduct subjective experiments on DIH99 dataset to to estimate the harmonization performance of our model on real-scenario dataset.

\textbf{RealHM real-scenario dataset.}
RealHM dataset\cite{jiang2021ssh} is a novel real-scenario dataset that satisfies the real image harmonization evaluation demands. It contains 216 high quality, high-resolution foreground/background image pairs with corresponding mask and harmonized outputs, where the foregrounds and the backgrounds include different human portraits and general objects and cover diverse environments. Different from the iHarmony4 synthetic dataset that perturbs the foreground objects by color-transfer methods~\cite{color_trans1,color_trans2}, the harmonized outputs in RealHM dataset are generated and adjusted by Photoshop tools and cover high-quality diverse environments. Hence, we additionally conduct qualitative comparisons on RealHM dataset to evaluate the harmonization performance of our model on real-world scenarios, while previous methods~\cite{dovenet,s2am,Ling_2021_CVPR,bargainet,Guo_2021_CVPR} do not take this evaluation.

\subsection{Evaluation Metrics}
\label{sec:eva}
\textbf{Objective Evaluation.} For evaluation, we validate our approaches on the iHarmony4 using Mean-Square-Errors~(MSE), Peak Signal-to-Noise Ratio (PSNR), Structural Similarity~(SSIM) as criteria metrics. Moreover,  we also report foreground MSE (fMSE) and foreground SSIM (fSSIM) to evaluate the harmonization performance only for foregroung regions, measuring how well the foreground is harmonized.

\begin{figure}[tbh!]
    \centering
    \includegraphics[width=\columnwidth]{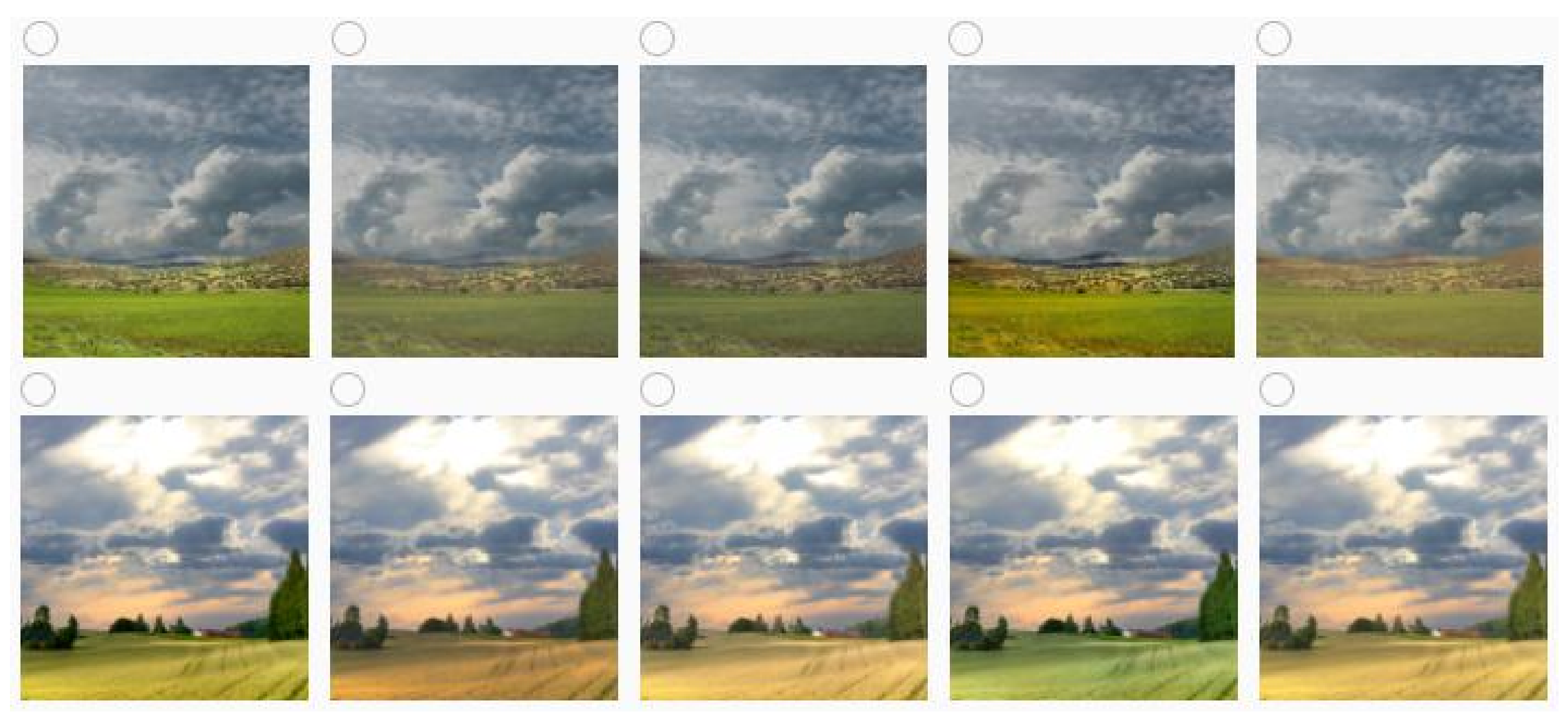}
    \caption{The layout of two image groups examples in our user study. The sequence in which the composite input and the harmonization results are displaying is randomly shuffled without annotations. }
    \label{fig:supp_user}
\end{figure}

\textbf{Subjective Evaluation.} To validate the generalization capacity in real-scenario applications, we evaluate our method on real-scenario dataset. Following \cite{dovenet}, \cite{Ling_2021_CVPR}, \cite{Guo_2021_CVPR}, we conduct subjective user study to compare our proposed method with current baseline methods on the DIH99 real composite dataset. In detail, we invite 28 participants from various majors to participate in subjective experiments. As shown in Figure~\ref{fig:supp_user}, each participant can access to a collection of image groups, each of which contains the original composite input and the harmonized results that are respectively generated by DIH\cite{DIH_Tsai}, BargainNet\cite{bargainet}, RainNet\cite{Ling_2021_CVPR} and our proposed method. To maintain unpredictability, we also shuffle the displaying order in each image group randomly. All subjects are not aware of image harmonization task, and are only required to choose the aesthetically superior image, contributing 28$\times$99 groups in total.

\begin{table*}[t!]
\centering
\caption{Quantitative comparisons on benchmark dataset iHarmony4~\cite{dovenet}. The best results are marked as \textbf{boldface}.}
\label{table:quantitative}
\begin{tabular}{@{}cl|c|cccccccccc@{}}
\toprule

\multicolumn{2}{c|}{} &  & \multicolumn{10}{c}{Sub-Dataset in iHarmony4}                                                                                                                                \\ \cmidrule(l){4-13} 
\multicolumn{2}{c|}{Methods}                         &      \#Param.↓                     & \multicolumn{2}{c|}{HCOCO}         & \multicolumn{2}{c|}{HAdobe5k}       & \multicolumn{2}{c|}{HFlickr}        & \multicolumn{2}{c|}{Hday2night}     & \multicolumn{2}{c}{All} \\
\multicolumn{2}{c|}{}                         &                           & MSE↓  & \multicolumn{1}{c|}{PSNR↑} & MSE↓   & \multicolumn{1}{c|}{PSNR↑} & MSE↓   & \multicolumn{1}{c|}{PSNR↑} & MSE↓   & \multicolumn{1}{c|}{PSNR↑} & MSE↓        & PSNR↑     \\ \midrule
\multicolumn{2}{c|}{input composite}          & -                         & 67.89 & \multicolumn{1}{c|}{34.07} & 342.27 & \multicolumn{1}{c|}{28.14} & 260.98 & \multicolumn{1}{c|}{28.35} & 107.95 & \multicolumn{1}{c|}{34.01} & 170.25      & 31.70      \\
\multicolumn{2}{c|}{Xue~\textit{et al.}~\cite{intro_illumination}}                & -                         & 77.04 & \multicolumn{1}{c|}{33.32} & 274.15 & \multicolumn{1}{c|}{28.79} & 249.54 & \multicolumn{1}{c|}{28.32} & 190.51 & \multicolumn{1}{c|}{31.24} & 155.87      & 31.40      \\
\multicolumn{2}{c|}{Zhu~\textit{et al.}~\cite{zhu2015learning}}                & -                         & 79.82 & \multicolumn{1}{c|}{33.04} & 414.31 & \multicolumn{1}{c|}{27.26} & 315.42 & \multicolumn{1}{c|}{27.52} & 136.71 & \multicolumn{1}{c|}{32.32} & 204.77      & 30.72     \\
\multicolumn{2}{c|}{DIH~\cite{DIH_Tsai}}                      & 41.76M                    & 51.85 & \multicolumn{1}{c|}{34.69} & 92.65  & \multicolumn{1}{c|}{32.28} & 163.38 & \multicolumn{1}{c|}{29.55} & 82.34  & \multicolumn{1}{c|}{34.62} & 76.77       & 33.41     \\
\multicolumn{2}{c|}{S$^2$AM\cite{s2am}}                     & 66.70M                    & 33.07 & \multicolumn{1}{c|}{36.09} & 48.22  & \multicolumn{1}{c|}{35.34} & 124.53 & \multicolumn{1}{c|}{31.00}    & 48.78  & \multicolumn{1}{c|}{35.60}  & 48.00          & 35.29     \\
\multicolumn{2}{c|}{DoveNet\cite{dovenet}}                  & 54.76M                    & 36.72 & \multicolumn{1}{c|}{35.83} & 52.32  & \multicolumn{1}{c|}{34.34} & 133.14 & \multicolumn{1}{c|}{30.21} & 54.05  & \multicolumn{1}{c|}{35.18} & 52.36       & 34.75     \\
\multicolumn{2}{c|}{BargainNet\cite{bargainet}}               & 58.74M                    & 24.84 & \multicolumn{1}{c|}{37.03} & 39.94  & \multicolumn{1}{c|}{35.34} & 97.32  & \multicolumn{1}{c|}{31.34} & 50.98  & \multicolumn{1}{c|}{35.67} & 37.82       & 35.88     \\
\multicolumn{2}{c|}{IIH\cite{Guo_2021_CVPR}}                      & 40.86M                    & 24.92 & \multicolumn{1}{c|}{37.16} & 43.02  & \multicolumn{1}{c|}{35.20}  & 105.13 & \multicolumn{1}{c|}{31.34} & 55.53  & \multicolumn{1}{c|}{35.96} & 38.71       & 35.90      \\
\multicolumn{2}{c|}{RainNet\cite{Ling_2021_CVPR}}                  & 54.75M                    & 31.12 & \multicolumn{1}{c|}{36.59} & 42.84  & \multicolumn{1}{c|}{36.20}  & 117.59 & \multicolumn{1}{c|}{31.33} & 47.24  & \multicolumn{1}{c|}{36.12} & 44.50        & 35.88     \\
\multicolumn{2}{c|}{Ours}                     & \textbf{28.37M}                    & \textbf{19.91} & \multicolumn{1}{c|}{\textbf{37.77}} & \textbf{21.16}  & \multicolumn{1}{c|}{\textbf{38.33}} & \textbf{66.14}  & \multicolumn{1}{c|}{\textbf{34.00}}    & \textbf{46.63}  & \multicolumn{1}{c|}{\textbf{37.11}} & \textbf{25.90}        & \textbf{37.50}      \\ \bottomrule
\end{tabular}
\end{table*}

\begin{table*}[t!]
\centering
\caption{Quantitative comparisons of foreground regions harmonization on benchmark dataset iHarmony4~\cite{dovenet}. The best results are marked as boldface.}
\label{table:fmse}
\begin{tabular}{@{}cl|cccccccccc@{}}
\toprule
\multicolumn{2}{c|}{} & \multicolumn{10}{c}{Sub-Dataset in   iHarmony4}                                                                                                                                                                                                                 \\ \cmidrule(l){3-12} 
\multicolumn{2}{c|}{Methods}                  & \multicolumn{2}{c|}{HCOCO}                            & \multicolumn{2}{c|}{HAdobe5k}                        & \multicolumn{2}{c|}{HFlickr}                          & \multicolumn{2}{c|}{Hday2night}                       & \multicolumn{2}{c}{All}          \\
\multicolumn{2}{c|}{}                  & fMSE↓           & \multicolumn{1}{c|}{fSSIM↑}         & fMSE↓          & \multicolumn{1}{c|}{fSSIM↑}         & fMSE↓           & \multicolumn{1}{c|}{fSSIM↑}         & fMSE↓           & \multicolumn{1}{c|}{fSSIM↑}         & fMSE↓           & fSSIM↑         \\ \midrule
\multicolumn{2}{c|}{input composite}   & 996.59          & \multicolumn{1}{c|}{82.57}          & 2051.61        & \multicolumn{1}{c|}{72.94}          & 1574.37         & \multicolumn{1}{c|}{80.31}          & 1409.98         & \multicolumn{1}{c|}{63.53}          & 1376.42         & 79.17          
\\

\multicolumn{2}{c|}{DIH\cite{DIH_Tsai}}              & 798.99          & \multicolumn{1}{c|}{82.25}          & 593.03         & \multicolumn{1}{c|}{77.77}           & 1099.13          & \multicolumn{1}{c|}{79.84}          & 1129.40          & \multicolumn{1}{c|}{62.77}          & 773.18          & 80.32 
\\
\multicolumn{2}{c|}{S$^2$AM~\cite{s2am}}              & 542.06          & \multicolumn{1}{c|}{84.77}          & 404.62         & \multicolumn{1}{c|}{81.20}           & 785.65          & \multicolumn{1}{c|}{82.33}          & 989.07          & \multicolumn{1}{c|}{63.74}          & 594.67          & 83.08          \\
\multicolumn{2}{c|}{DoveNet~\cite{dovenet}}           & 551.01          & \multicolumn{1}{c|}{84.73}          & 380.36         & \multicolumn{1}{c|}{83.09}          & 827.03          & \multicolumn{1}{c|}{82.35}          & 1075.71         & \multicolumn{1}{c|}{61.94}          & 532.62          & 83.57          \\
\multicolumn{2}{c|}{IIH\cite{Guo_2021_CVPR}}               & 416.38          & \multicolumn{1}{c|}{86.19}          & 284.21         & \multicolumn{1}{c|}{83.64}          & 716.60           & \multicolumn{1}{c|}{82.97}          & 797.04          & \multicolumn{1}{c|}{64.49}          & 400.29          & 84.69          \\

\multicolumn{2}{c|}{Ours}     & \textbf{300.19} & \multicolumn{1}{c|}{\textbf{89.81}} & \textbf{157.40} & \multicolumn{1}{c|}{\textbf{92.54}} & \textbf{412.82} & \multicolumn{1}{c|}{\textbf{91.99}} & \textbf{763.66} & \multicolumn{1}{c|}{\textbf{77.79}} & \textbf{279.46} & \textbf{90.69} \\ \bottomrule
\end{tabular}
\end{table*}

\begin{table*}[tbh!]
\centering
\caption{Quantitative comparisons of harmonizing image at various resolutions on HAdobe5k dataset.}
\label{table:hr}
\begin{tabular}{@{}c|c|ccccccc@{}}
\toprule
Dataset                   & Resolution & S$^2$AM~\cite{s2am}  & DoveNet~\cite{dovenet} & BargainNet~\cite{bargainet} & RainNet\cite{Ling_2021_CVPR} & IIH~\cite{Guo_2021_CVPR}   & S$^2$CRNet-VGG~\cite{s2crnet} & Ours  \\ \midrule
                            & 256×256    & 35.34 & 34.34   & 35.34      & 35.77   & 35.26 & 36.42   & \textbf{37.50}  \\
\multicolumn{1}{c|}{HAdobe5k} & 512×512    & 32.55 & 34.72   & 35.78      & 36.56   & 36.62 & 36.89   & \textbf{37.50}  \\
                            & 1024×1024  & 30.02 & 34.43   & 34.64      & 36.81   & 36.45 & 36.52   & \textbf{37.55} \\ \bottomrule
\end{tabular}
\end{table*}

\begin{figure*}[t!]
    \centering
    \includegraphics[width=\textwidth]{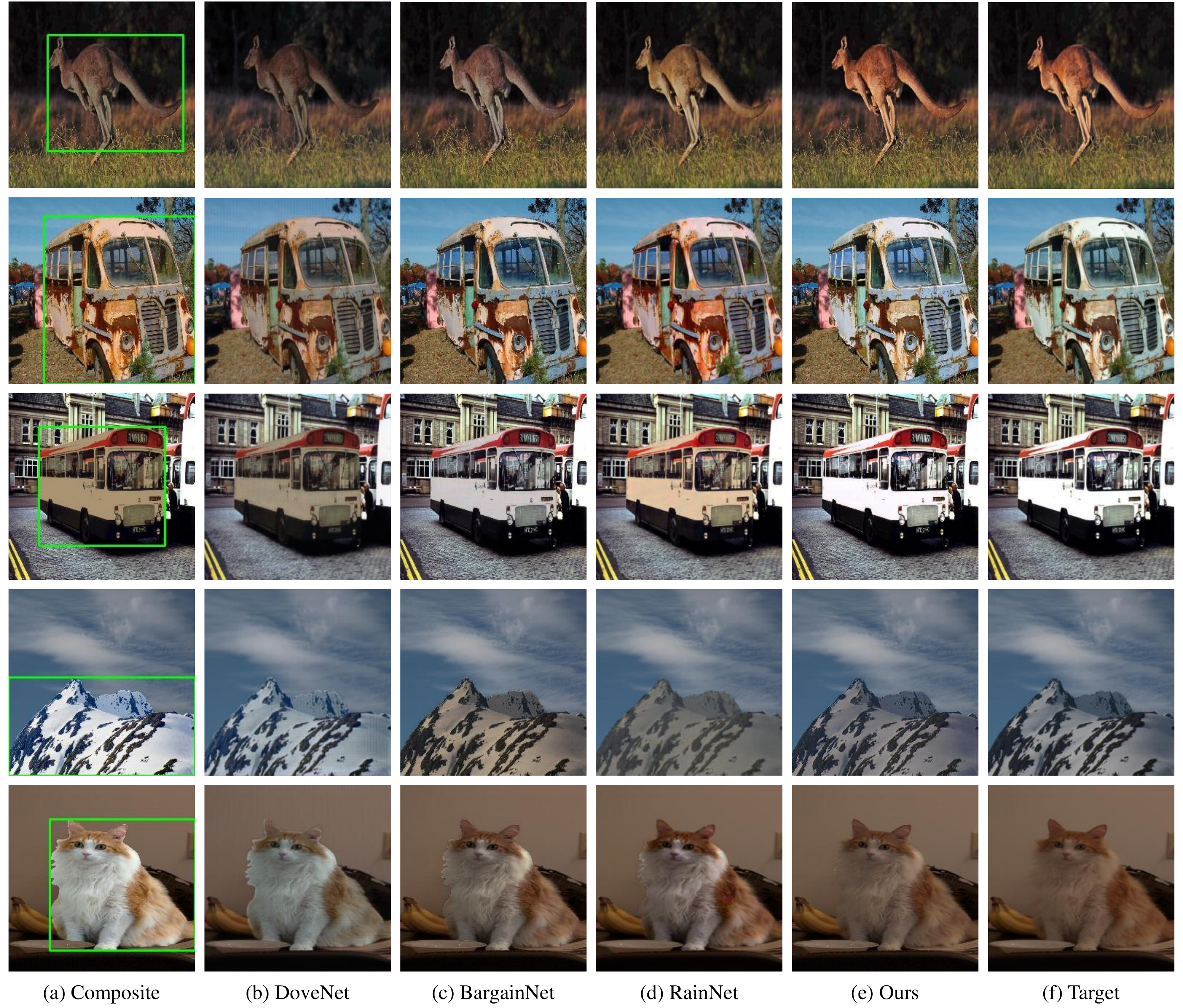}
    \caption{Qualitative comparisons with other methods on iHarmony4 Dataset.  Foreground regions are marked by green boxes in composite images. }
    \label{fig:compare_all}
\end{figure*}

\subsection{Comparisons with State-of-the-art Methods}
We compare our method with current solid image harmonization algorithms~(\cite{intro_illumination}, \cite{zhu2015learning}, DIH~\cite{DIH_Tsai}, S$^2$AM~\cite{s2am},  DoveNet~\cite{dovenet}, BargainNet~\cite{bargainet}, RainNet~\cite{Ling_2021_CVPR} and IIH~\cite{Guo_2021_CVPR}). 

 In Table~\ref{table:quantitative} and Table~\ref{table:fmse}, we report the quantitative comparisons of the harmonization performance across four sub-datasets of iHarmony4 , where all the previous methods employ their official configurations and pre-trained models. All our experiments are trained and evaluated on the iHarmony4 dataset and our framework is optimized under the loss function in Eq~\ref{eq:loss}. All the numeric  metrics are calculated on the testing datasets in iHarmony4 and we additionally report the model size of each method. 
 
 From Table~\ref{table:quantitative}, even with one-third fewer parameters, our approach substantially outperforms current solid baselines by a large margin at PSNR and MSE metrics, with 12.79 reduction at overall MSE metric and 1.6 improvement at overall PSNR metric. Additionally, we evaluate the foreground harmonization quality following \cite{Guo_2021_CVPR}. All quantitative results are collected from \cite{Guo_2021_CVPR} and we compare them with our method in Table~\ref{table:fmse}. Specifically for the foreground harmonization, our approach outperforms previous methods in terms of the pixel-level harmonization~(120.83 reduction at fMSE metric) and the structure-level similarity harmonization~(6.00 improvement at fSSIM metric).
 
In addition to the quantitative comparisons, we also present some qualitative comparisons in Figure~\ref{fig:compare_all}. Our method achieves better harmonization quality compared to~\cite{dovenet}, \cite{bargainet}, \cite{Ling_2021_CVPR}. Significantly, we find that our method not only performs well in harmonizing the composite foreground region according to the background appearance, but also has strong reconstruction ability for those foreground regions that have unusual styles, from which it also infers the limitations of previous methods.

 Furthermore, we attempt to investigate the model performance in harmonizing higher resolution images. Following~\cite{s2crnet}, the high-resolution images in HAdobe5k dataset are first re-scaled into $256\times256, 512\times512, 1024\times1024$ resolutions, and we evaluate the harmonization performance using the PSNR metric at different resolutions. We compare the quantitative performance with current low-resolution harmonization methods~\cite{s2am}, \cite{dovenet}, \cite{bargainet}, \cite{Ling_2021_CVPR}, \cite{Guo_2021_CVPR} and high-resolution harmonization method \cite{s2crnet} using their default pre-trained settings,and we report the numeric results in Table~\ref{table:hr}. In Table~\ref{table:hr}, we find that the previous methods for low resolution harmonization confront a performance decline as the image resolution rises. We conjecture that the UNet-based structures show limitations in processing large scale feature map as the number of pixels that will present the vital details is significantly reduced. Particularly, those methods which approach to leverage global information~(RAIN block in~\cite{Ling_2021_CVPR}, global rendering curves in~\cite{s2crnet} and our proposed style fusion) obtain more stable or even better performance when harmonizing high-resolution images, where our method maintains the higher performance than others.

\begin{figure*}[tbh!]
    \centering
    \includegraphics[width=0.85\textwidth]{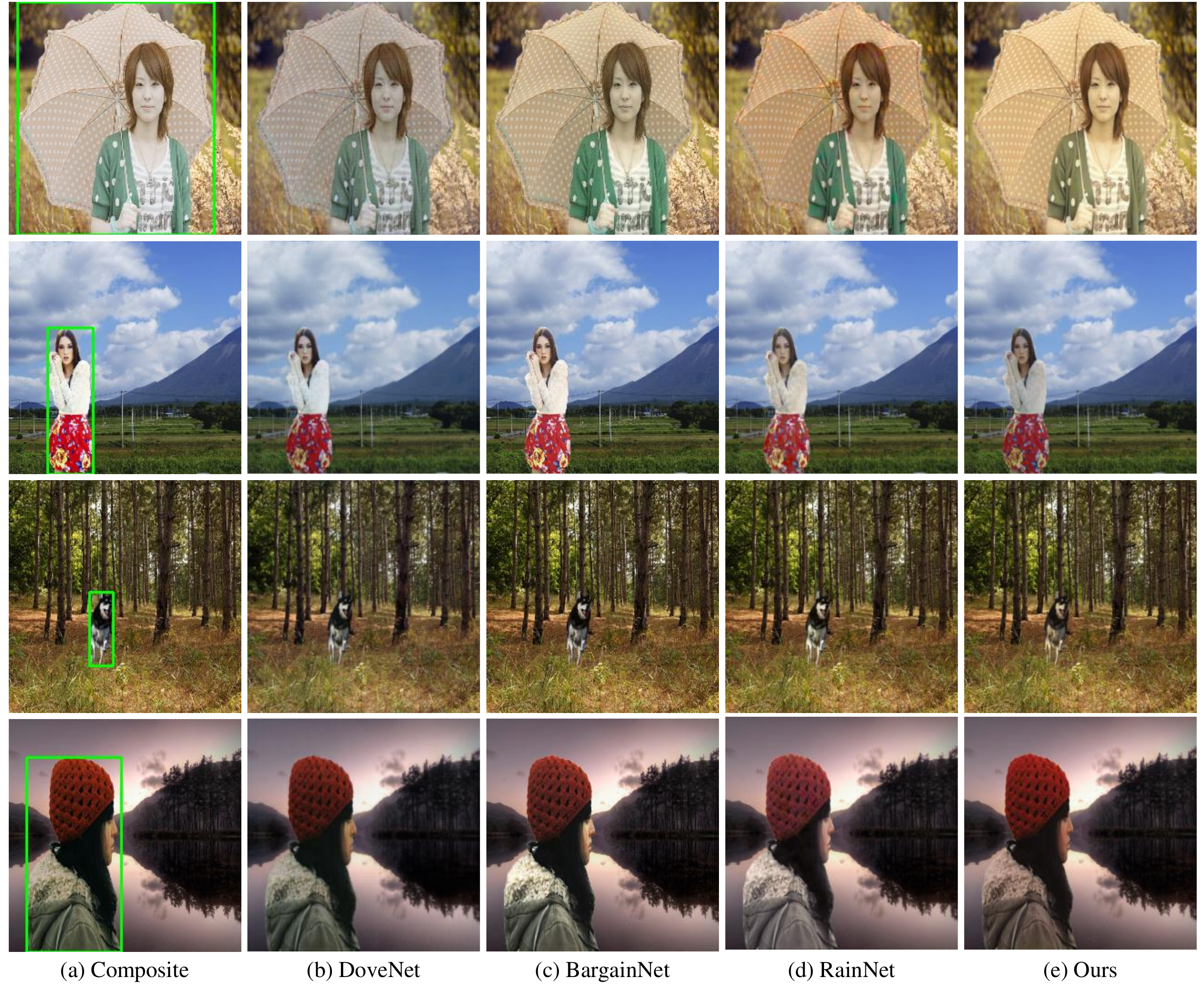}
    \caption{Visual harmonization comparisons on real-scenario datasets DIH99\cite{DIH_Tsai}. Foreground regions are marked by green boxes in composite images.}
    \label{fig:compare_dih}
\end{figure*}

\begin{figure*}[tbh!]
    \centering
    \includegraphics[width=0.9\textwidth]{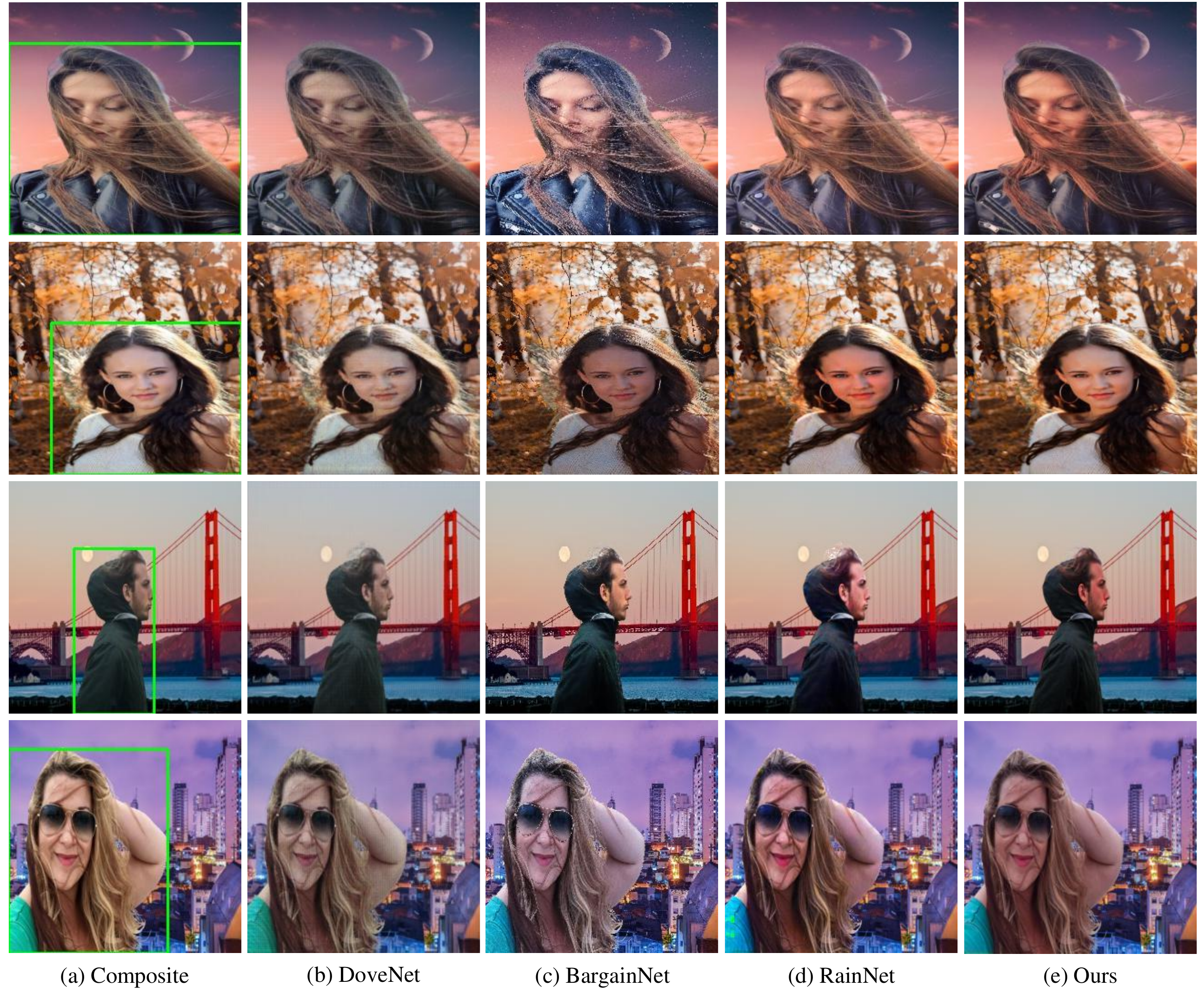}
    \caption{Visual harmonization comparisons on real-scenario datasets RealHM~\cite{jiang2021ssh}.  Foreground regions are marked by green boxes in composite images.}
    \label{fig:compare_realhm}
\end{figure*}

\subsection{Performance on Real-scenario Datasets.}
\begin{table}[t!]
\centering
\caption{Voting results of the user study on DIH99~\cite{DIH_Tsai} real-scenario dataset.}
\begin{tabular}{@{}c|ccccc@{}}
\toprule
Method     & Copy-paste  & DIH         & BargainNe  & RainNet     & Ours                 \\ \midrule
Votes      & 326         & 521         & 491         & 618         & \textbf{816}         \\
Preference & 11.76\% & 18.80\% & 17.71\% & 22.29\% & \textbf{29.44\%} \\ \bottomrule
\end{tabular}
\label{table:user_study}
\end{table}


In addition to investigate the generalization ability in real-scenario applications, we evaluate our method on real-scenario datasets, DIH~\cite{DIH_Tsai} and RealHM~\cite{jiang2021ssh} and present some visualized harmonization results in Figure~\ref{fig:compare_dih} and Figure~\ref{fig:compare_realhm}. Our method obtains more favorable results in harmonizing real-scenario copy-paste image compared to previous methods. From the visualized results in Figure~\ref{fig:compare_dih} and Figure~\ref{fig:compare_realhm}, we could see that our method also show advantages in generating more harmonious foregrounds that share similar style~(hue, color, brightness, $etc.$) with the background images.   

Moreover, we conduct subjective user study to compare our proposed method with baseline methods~(original copy-paste image, DIH~\cite{DIH_Tsai}, BargainNet~\cite{bargainet} and RainNet~\cite{Ling_2021_CVPR} on  the DIH99 real composite dataset. The details of the user study is introduced in Section~\ref{sec:eva}. We summarize the total voting times of each method and calculate the preference ranking in Table~\ref{table:user_study}. Our method receives the most votes in the subjective user study. The qualitative comparisons and the subjective user study demonstrate that our method has good generalization capacity in real-world scenarios. 


\begin{figure*}[tbh!]
    \centering
    \begin{adjustbox}{width=0.9\textwidth}
    \includegraphics[width=\textwidth]{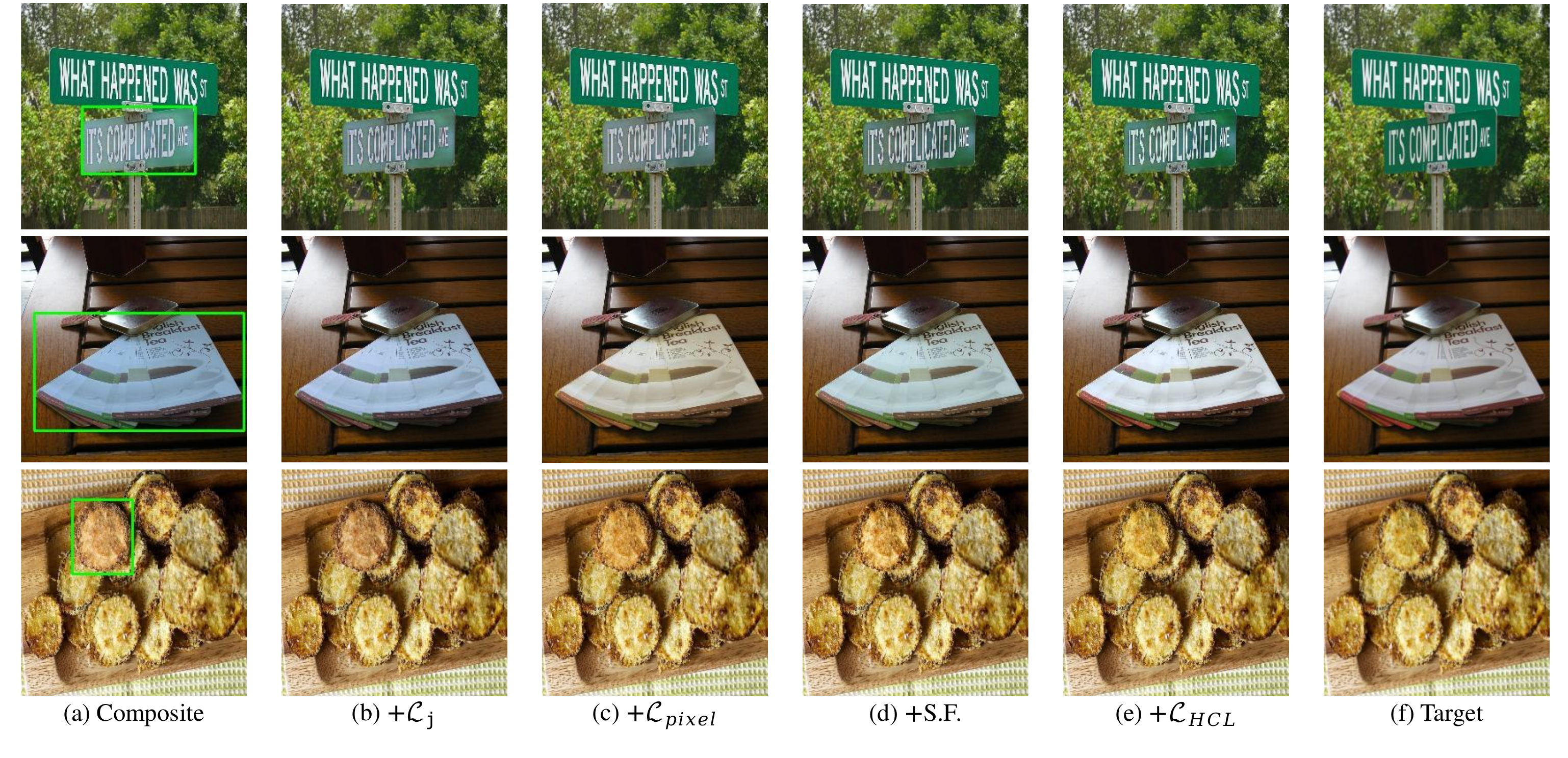}
    \end{adjustbox}
    \caption{Visualized examples in ablation studies. S.F. denotes our proposed style fusion block. }
    \label{fig:abl_example}
\end{figure*}

\subsection{Ablation Study}
In this section, we conduct the ablation experiments to investigate the effectiveness of each component in our method, including the proposed style fusion strategy and the employed loss functions. 


\textbf{Effectiveness of External Style Fusion.}
As we introduce that we add the background style guidance from the encoder to the decoder via the proposed style fusion, we first investigate its harmonization performance while only using the basic loss function $\mathcal{L}_{1}$ and $\mathcal{L}_{pixel}$. From the model \emph{B} and \emph{C} in Table~\ref{table:ablation}, we find that our designed style fusion improves the baseline model by a large margin. It shows that adding external background guidance not only performs well in pixel level~(PSNR, MSE and fMSE)  but also structure level~(SSIM and fSSIM) harmonization, which exactly proves the effectiveness of our proposed style fusion.

\begin{table}[tbh!]
\centering
\caption{Ablation studies. S.F. denotes our proposed style fusion strategy.}
\begin{tabular}{@{}c|cccc|ccc@{}}
\toprule
\# & $\mathcal{L}_{1}$ & $\mathcal{L}_{pixel}$ & S.F. & $\mathcal{L}_{H C L}$ & MSE~$\downarrow$            & PSNR~$\uparrow$          & fSSIM~$\uparrow$    \\ \midrule
1  & $\checkmark$  &     &    &     & 54.55          & 34.56         & 85.68  \\
2  & $\checkmark$  & $\checkmark$   &    &     & 42.14          & 35.53         & 88.14  \\
3  & $\checkmark$  & $\checkmark$   & $\checkmark$  &     & 29.08          & 37.01         & 90.03\\
4  & $\checkmark$  & $\checkmark$   & $\checkmark$  & $\checkmark$   & \textbf{25.92} & \textbf{37.50} & \textbf{90.69} \\ \bottomrule
\end{tabular}
\label{table:ablation}
\end{table}

\begin{figure}[t]
    \centering

    \includegraphics[width=\columnwidth]{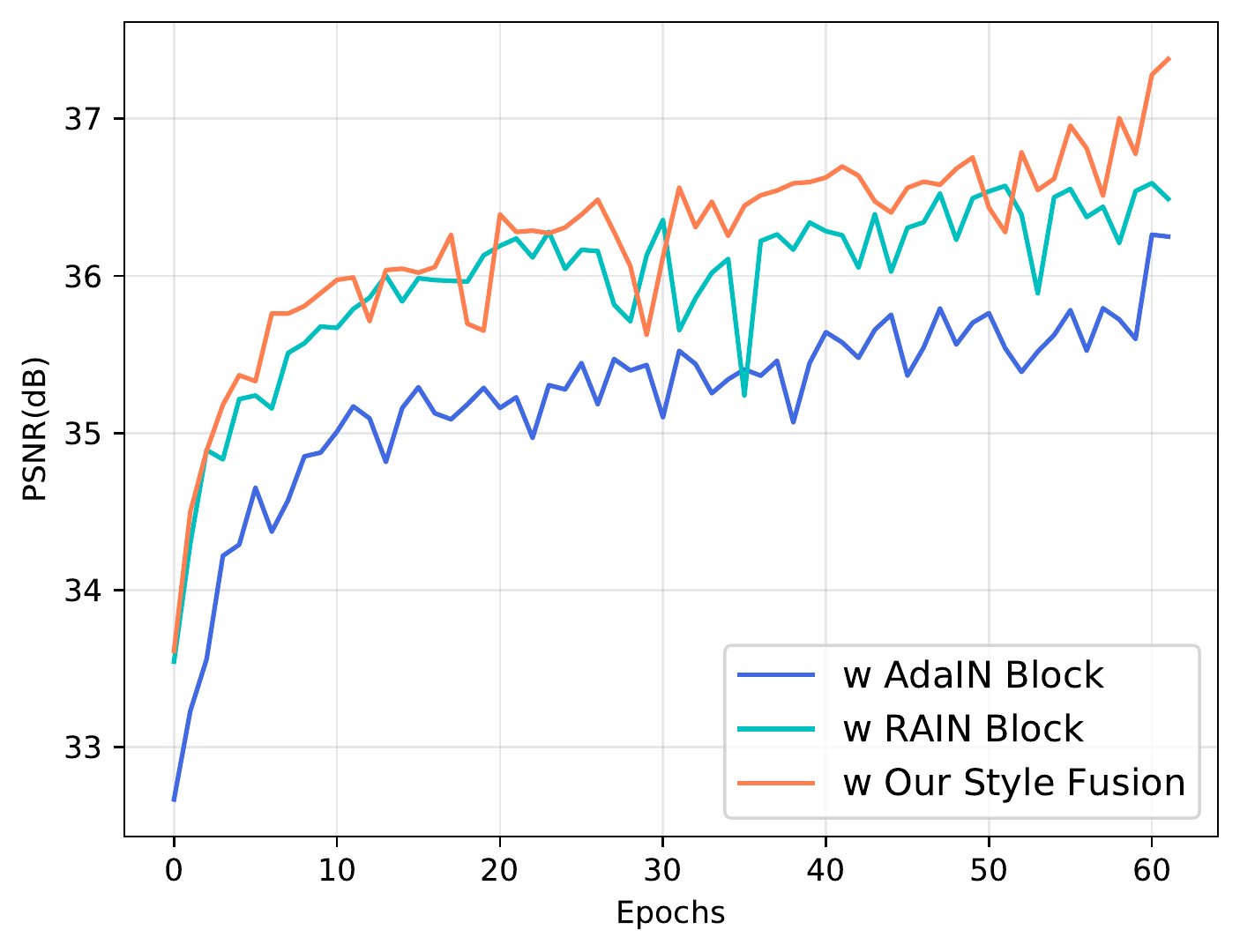}
    \caption{Comparisons of inserting AdaIN block~\cite{adain}, RAIN block~\cite{Ling_2021_CVPR} and the proposed style fusion in our framework at PSNR metric. The higher PSNR score denotes the better composite image has been harmonized. }
    \label{fig:fusion_psnr}
\end{figure}

\begin{figure}[t]
    \centering
    \includegraphics[width=\columnwidth]{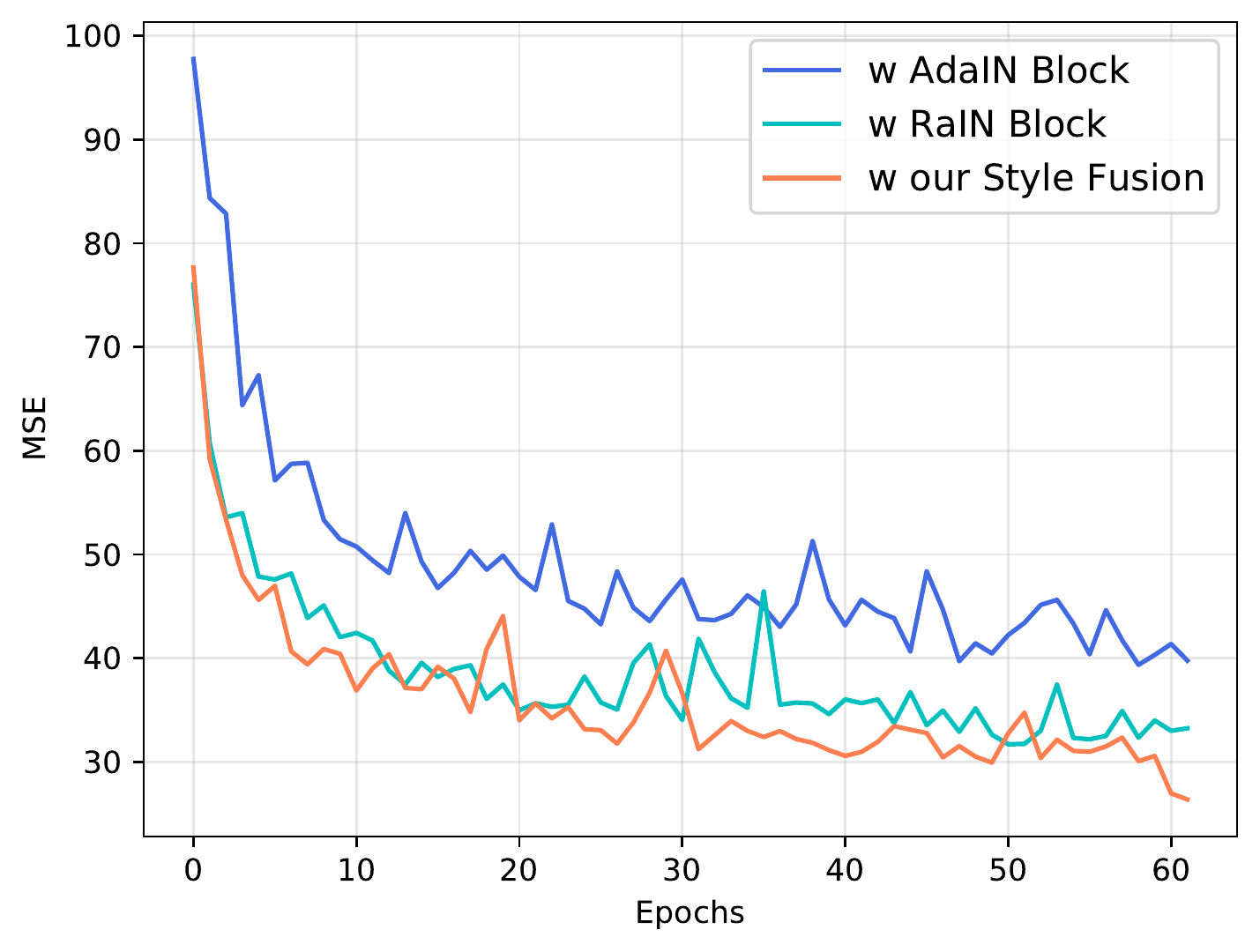}
    \caption{Comparisons of inserting AdaIN block~\cite{adain}, RAIN block~\cite{Ling_2021_CVPR} and the proposed style fusion in our framework at MSE metric. The lower MSE score denotes the better composite image has been harmonized. }
    \label{fig:fusion_mse}
\end{figure}

\begin{figure}[t]
    \centering
    \includegraphics[width=0.5\textwidth]{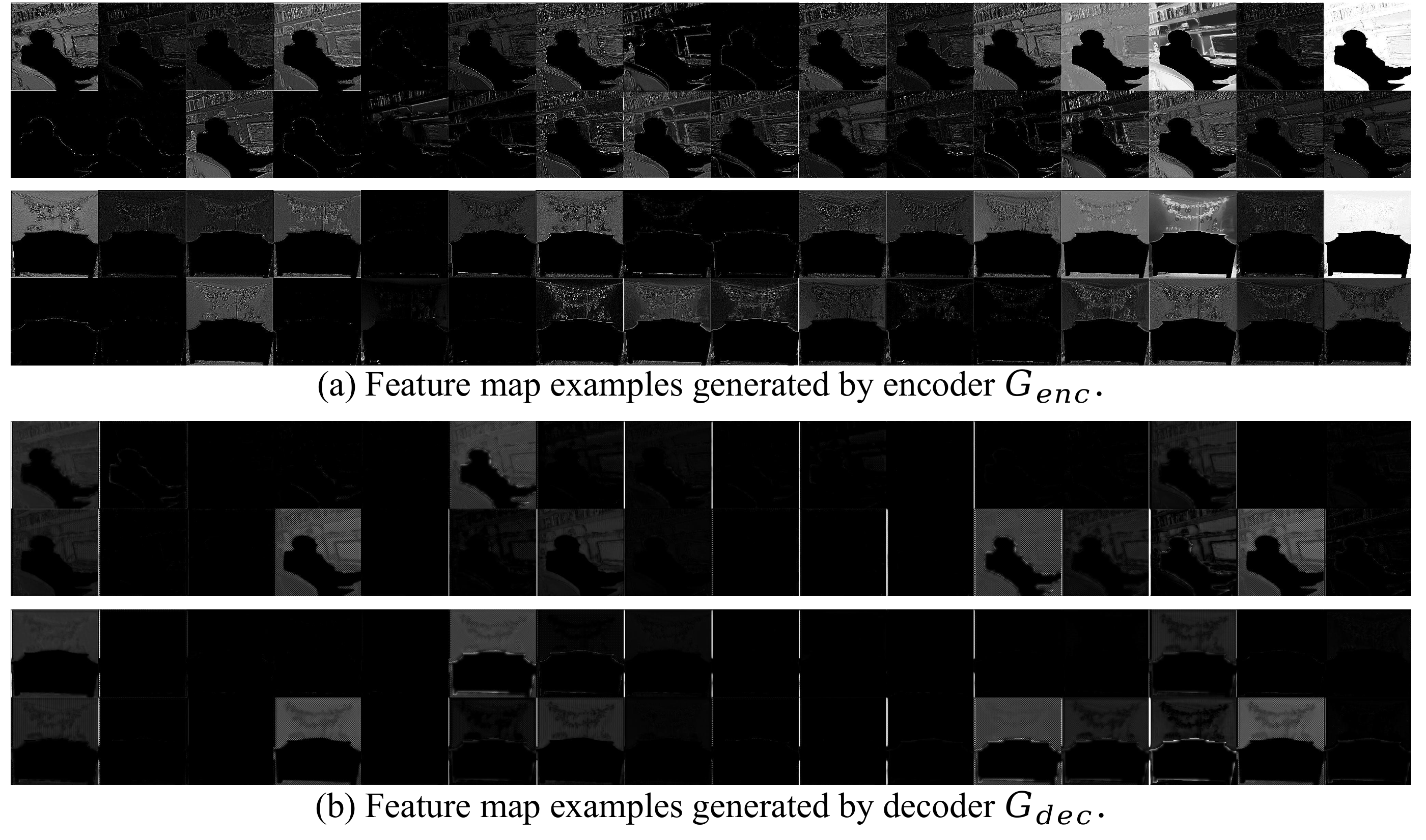}
    \caption{Background feature map examples in $G_{enc}$ and $G_{dec}$. Notice that only the feature maps of background regions are shown for comparison. As seen in the examples,  background features contain more details when extracting from the encoder.}
    \label{fig:supp_feature}
\end{figure}

To further investigate its effectiveness, we compare our style fusion with AadIN~\cite{adain} and RainNet~\cite{Ling_2021_CVPR}, which apply adaptive instance normalization to the internal feature maps within the decoder. Specifically, we replace all of the style fusion layers in our framework with the AadIN blocks and RAIN blocks respectively, and preliminarily train for 60 epochs. We quantify the harmonization performance by PSNR metric and MSE metric, and report the validation results in Figure~\ref{fig:fusion_psnr} and Figure~\ref{fig:fusion_mse} respectively. When using style fusion instead of RAIN block during the entire training phrase, the model performance is considerably enhanced. Additionally in Figure~\ref{fig:supp_feature}, we present two groups of feature map examples that generated by encoder $G_{enc}$ and decoder $G_{dec}$, respectively.
Because our style fusion relies on the encoder guidance, the comparisons also proves our proposition that the unmodified background features convey more accurate region-specific details when extracting from the encoder. The foregoing experiments effectively validate the proposed style fusion, making it useful and indispensable in our image harmonization task.    


\textbf{Effectiveness of Contrastive Loss Function.}
From model \emph{A} and \emph{B} in Table~\ref{table:ablation}, we find that employing $\mathcal{L}_{pixel}$ rather than only comparing the whole image with $\mathcal{L}_{1}$ helps to improve the harmonization performance. From the model \emph{B} to \emph{D} in Table~\ref{table:ablation}, the proposed contrastive loss further improve the harmonization quality after applying our style fusion. We present a visualized example in the ablation studies in Figure~\ref{fig:abl_example} and we find that although the style fusion achieves preliminary harmonization, the designed $\mathcal{L}_{HCL}$ makes the foreground car more compatible with the background region.  We conjecture that the importance of using our region-wise contrastive loss is due to the effective comparison between the background and foreground style features, making the reconstructed foreground more harmonious. Moreover, as our method is set as supervised scheme, the ground-truth background $\widehat{H}^{bg}$ provides more stable feature representations than the harmonized background $H^{bg}$.

\textbf{The Quantities of Sampling Patches.}
Table~\ref{table:ablation_k} compares the performance with different patch sampling numbers~\emph{K} during the sample generation~(Section 3.3.1). Although increasing sampling numbers theoretically improves contrast complexity between foreground and background style features, the ablation results reveal that capturing more than 256 patches degrades harmonization performance with a modest drop at PSNR and SSIM metrics.
We conjecture that as the number of negative samples grows, more complex style representations emerge, and our method does not employ a well-designed feature extractor to obtain unified style representations following~\cite{patchnce} but only uses a shared-~\emph{MLP} to obtain embedded positive and negative vectors~(illustrated in Section 3.3.1). Moreover, when cropping and embedding more patches, the training resources also increase, such as computational cost and processing time. Overall, we set $K=256$ for a trade-off between the harmonization performance and the training complexity.   

\begin{table}[t!]
\centering
\caption{Ablation studies of patch number \emph{K}.}
\begin{tabular}{@{}c|ccc@{}}
\toprule
Patches of \emph{K}    & MSE↓          & PSNR↑         & SSIM↑          \\ \midrule
128  & 29.06         & 37.00            & 98.92          \\
256  & \textbf{25.90} & \textbf{37.50} & \textbf{99.04} \\
512  & 30.87         & 36.79         & 98.88          \\
1024 & 30.35         & 36.69         & 98.83          \\ \bottomrule
\end{tabular}
\label{table:ablation_k}
\end{table}

\subsection{Limitations and Social Impacts.}    
Our proposed method achieves state-of-the-art performance specifically addressing the image harmonization task. However, we have not generalized our approaches to other harmonization models and we conduct harmonization only on low-resolution images as previous methods do. We would like to investigate the robustness of the proposed external style fusion and region-wise contrastive loss in more generalized applications for future work. Meanwhile, we notice that there are several potential negative impacts caused by image harmonization techniques. As our task aims to harmonized composite image, some may use it to create fake multimedia information and false data in academic papers, which may not be structurally detected by some effective forgery detection methods~\cite{zhou2018learning, foregy1, foregy2}.

\section{Conclusion}
\label{sec:conclusion}
In this paper, we design a novel region-wise contrastive loss function for image harmonization to explore the information from both foreground and background regions as negative and positive, respectively. With the proposed contrastive learning scheme, it enables our method to bring together harmonized foreground styles and ground-truth background styles by regularizing the negative samples to be more relative to the positive samples.
Furthermore,  while previous works focus on reconstructing the foreground with background styles inside the generator, we design an efficient style fusion strategy that employs external background style representations as reference to generate more harmonized image. The external background representations are first extracted from the encoder of UNet-based structure and work as guidance conspired with adaptive Instance Normalization in each layer of the generator. 
The extensive experiments demonstrate the feasibility and the effectiveness of our method, which also maintains favorable generalization ability on real-scenario applications.

\bibliographystyle{IEEEtran}
\bibliography{ref.bib}

\begin{thebibliography}{10}
\providecommand{\url}[1]{#1}
\csname url@samestyle\endcsname
\providecommand{\newblock}{\relax}
\providecommand{\bibinfo}[2]{#2}
\providecommand{\BIBentrySTDinterwordspacing}{\spaceskip=0pt\relax}
\providecommand{\BIBentryALTinterwordstretchfactor}{4}
\providecommand{\BIBentryALTinterwordspacing}{\spaceskip=\fontdimen2\font plus
\BIBentryALTinterwordstretchfactor\fontdimen3\font minus
  \fontdimen4\font\relax}
\providecommand{\BIBforeignlanguage}[2]{{%
\expandafter\ifx\csname l@#1\endcsname\relax
\typeout{** WARNING: IEEEtran.bst: No hyphenation pattern has been}%
\typeout{** loaded for the language `#1'. Using the pattern for}%
\typeout{** the default language instead.}%
\else
\language=\csname l@#1\endcsname
\fi
#2}}
\providecommand{\BIBdecl}{\relax}
\BIBdecl

\bibitem{editing1}
J.~Zhu, Y.~Shen, D.~Zhao, and B.~Zhou, ``In-domain gan inversion for real image
  editing,'' in \emph{ECCV}.\hskip 1em plus 0.5em minus 0.4em\relax Springer,
  2020, pp. 592--608.

\bibitem{encoder3}
T.~Wang, Y.~Zhang, Y.~Fan, J.~Wang, and Q.~Chen, ``High-fidelity gan inversion
  for image attribute editing,'' \emph{arXiv preprint arXiv:2109.06590}, 2021.

\bibitem{relighting}
H.~Zhou, S.~Hadap, K.~Sunkavalli, and D.~W. Jacobs, ``Deep single portrait
  image relighting,'' in \emph{ICCV}, 2019.

\bibitem{enhancement}
M.~Gharbi, J.~Chen, J.~T. Barron, S.~W. Hasinoff, and F.~Durand, ``Deep
  bilateral learning for real-time image enhancement,'' \emph{TOG}, vol.~36,
  no.~4, pp. 1--12, 2017.

\bibitem{synthesis1}
J.~Cao, Y.~Hu, B.~Yu, R.~He, and Z.~Sun, ``3d aided duet gans for multi-view
  face image synthesis,'' \emph{TIFS}, vol.~14, no.~8, pp. 2028--2042, 2019.

\bibitem{synthesis3}
Y.~Sun, J.~Tang, X.~Shu, Z.~Sun, and M.~Tistarelli, ``Facial age synthesis with
  label distribution-guided generative adversarial network,'' \emph{TIFS},
  vol.~15, pp. 2679--2691, 2020.

\bibitem{synthesis2}
Y.~Shi, X.~Yang, Y.~Wan, and X.~Shen, ``Semanticstylegan: Learning
  compositional generative priors for controllable image synthesis and
  editing,'' in \emph{CVPR}, 2022, pp. 11\,254--11\,264.

\bibitem{foregy1}
T.~J. De~Carvalho, C.~Riess, E.~Angelopoulou, H.~Pedrini, and
  A.~de~Rezende~Rocha, ``Exposing digital image forgeries by illumination color
  classification,'' \emph{TIFS}, vol.~8, no.~7, pp. 1182--1194, 2013.

\bibitem{foregy2}
Y.~Liu, X.~Zhu, X.~Zhao, and Y.~Cao, ``Adversarial learning for constrained
  image splicing detection and localization based on atrous convolution,''
  \emph{TIFS}, vol.~14, no.~10, pp. 2551--2566, 2019.

\bibitem{forgery3}
L.~Zhuo, S.~Tan, B.~Li, and J.~Huang, ``Self-adversarial training incorporating
  forgery attention for image forgery localization,'' \emph{TIFS}, vol.~17, pp.
  819--834, 2022.

\bibitem{video_syn}
T.-C. Wang, M.-Y. Liu, J.-Y. Zhu, G.~Liu, A.~Tao, J.~Kautz, and B.~Catanzaro,
  ``Video-to-video synthesis,'' in \emph{NIPS}, 2018.

\bibitem{lee2019inserting}
D.~Lee, T.~Pfister, and M.-H. Yang, ``Inserting videos into videos,'' in
  \emph{CVPR}, 2019, pp. 10\,061--10\,070.

\bibitem{intro_color1}
J.-F. Lalonde and A.~A. Efros, ``Using color compatibility for assessing image
  realism,'' in \emph{ICCV}.\hskip 1em plus 0.5em minus 0.4em\relax IEEE, 2007,
  pp. 1--8.

\bibitem{intro_color2}
E.~Reinhard, M.~Adhikhmin, B.~Gooch, and P.~Shirley, ``Color transfer between
  images,'' \emph{CG\&A}, vol.~21, no.~5, pp. 34--41, 2001.

\bibitem{intro_illumination}
S.~Xue, A.~Agarwala, J.~Dorsey, and H.~Rushmeier, ``Understanding and improving
  the realism of image composites,'' \emph{TOG}, vol.~31, no.~4, pp. 1--10,
  2012.

\bibitem{intro_texture}
K.~Sunkavalli, M.~K. Johnson, W.~Matusik, and H.~Pfister, ``Multi-scale image
  harmonization,'' \emph{TOG}, vol.~29, no.~4, pp. 1--10, 2010.

\bibitem{s2am}
X.~Cun and C.-M. Pun, ``Improving the harmony of the composite image by
  spatial-separated attention module,'' \emph{TIP}, vol.~29, pp. 4759--4771,
  2020.

\bibitem{dovenet}
W.~Cong, J.~Zhang, L.~Niu, L.~Liu, Z.~Ling, W.~Li, and L.~Zhang, ``Dovenet:
  Deep image harmonization via domain verification,'' in \emph{CVPR}, 2020, pp.
  8394--8403.

\bibitem{bargainet}
W.~Cong, L.~Niu, J.~Zhang, J.~Liang, and L.~Zhang, ``{BargainNet}:
  Background-guided domain translation for image harmonization,'' in
  \emph{ICME}, 2021.

\bibitem{Ling_2021_CVPR}
J.~Ling, H.~Xue, L.~Song, R.~Xie, and X.~Gu, ``Region-aware adaptive instance
  normalization for image harmonization,'' in \emph{CVPR}, June 2021, pp.
  9361--9370.

\bibitem{unet}
O.~Ronneberger, P.~Fischer, and T.~Brox, ``U-net: Convolutional networks for
  biomedical image segmentation,'' in \emph{MICCAI}.\hskip 1em plus 0.5em minus
  0.4em\relax Springer, 2015, pp. 234--241.

\bibitem{infonce}
A.~Van~den Oord, Y.~Li, and O.~Vinyals, ``Representation learning with
  contrastive predictive coding,'' \emph{arXiv e-prints}, pp. arXiv--1807,
  2018.

\bibitem{encoder1}
O.~Tov, Y.~Alaluf, Y.~Nitzan, O.~Patashnik, and D.~Cohen-Or, ``Designing an
  encoder for stylegan image manipulation,'' \emph{TOG}, vol.~40, no.~4, pp.
  1--14, 2021.

\bibitem{encoder2}
R.~Abdal, Y.~Qin, and P.~Wonka, ``Image2stylegan: How to embed images into the
  stylegan latent space?'' in \emph{ICCV}, 2019, pp. 4432--4441.

\bibitem{rw_color1}
D.~Cohen-Or, O.~Sorkine, R.~Gal, T.~Leyvand, and Y.-Q. Xu, ``Color
  harmonization,'' in \emph{SIGGRAPH}, 2006, pp. 624--630.

\bibitem{rw_color2}
V.~Bychkovsky, S.~Paris, E.~Chan, and F.~Durand, ``Learning photographic global
  tonal adjustment with a database of input/output image pairs,'' in
  \emph{CVPR}.\hskip 1em plus 0.5em minus 0.4em\relax IEEE, 2011, pp. 97--104.

\bibitem{rw_gradient1}
J.~Jia, J.~Sun, C.-K. Tang, and H.-Y. Shum, ``Drag-and-drop pasting,''
  \emph{TOG}, vol.~25, no.~3, pp. 631--637, 2006.

\bibitem{rw_gradient2}
P.~P{\'e}rez, M.~Gangnet, and A.~Blake, ``Poisson image editing,'' in
  \emph{SIGGRAPH}, 2003, pp. 313--318.

\bibitem{zhu2015learning}
J.-Y. Zhu, P.~Krahenbuhl, E.~Shechtman, and A.~A. Efros, ``Learning a
  discriminative model for the perception of realism in composite images,'' in
  \emph{ICCV}, 2015, pp. 3943--3951.

\bibitem{DIH_Tsai}
Y.-H. Tsai, X.~Shen, Z.~Lin, K.~Sunkavalli, X.~Lu, and M.-H. Yang, ``Deep image
  harmonization,'' in \emph{CVPR}, 2017.

\bibitem{sofiiuk2021foreground}
K.~Sofiiuk, P.~Popenova, and A.~Konushin, ``Foreground-aware semantic
  representations for image harmonization,'' in \emph{WACV}, 2021, pp.
  1620--1629.

\bibitem{Guo_2021_CVPR}
Z.~Guo, H.~Zheng, Y.~Jiang, Z.~Gu, and B.~Zheng, ``Intrinsic image
  harmonization,'' in \emph{CVPR}, June 2021, pp. 16\,367--16\,376.

\bibitem{ihtrans}
Z.~Guo, D.~Guo, H.~Zheng, Z.~Gu, B.~Zheng, and J.~Dong, ``Image harmonization
  with transformer,'' in \emph{ICCV}, 2021, pp. 14\,870--14\,879.

\bibitem{contrastive1}
R.~Hadsell, S.~Chopra, and Y.~LeCun, ``Dimensionality reduction by learning an
  invariant mapping,'' in \emph{CVPR}, vol.~2.\hskip 1em plus 0.5em minus
  0.4em\relax IEEE, 2006, pp. 1735--1742.

\bibitem{moco}
K.~He, H.~Fan, Y.~Wu, S.~Xie, and R.~Girshick, ``Momentum contrast for
  unsupervised visual representation learning,'' in \emph{CVPR}, 2020, pp.
  9729--9738.

\bibitem{byol}
J.-B. Grill, F.~Strub, F.~Altch{\'e}, C.~Tallec, P.~Richemond, E.~Buchatskaya,
  C.~Doersch, B.~Avila~Pires, Z.~Guo, M.~Gheshlaghi~Azar \emph{et~al.},
  ``Bootstrap your own latent-a new approach to self-supervised learning,''
  \emph{NIPS}, vol.~33, pp. 21\,271--21\,284, 2020.

\bibitem{simclr}
T.~Chen, S.~Kornblith, M.~Norouzi, and G.~Hinton, ``A simple framework for
  contrastive learning of visual representations,'' in \emph{ICML}.\hskip 1em
  plus 0.5em minus 0.4em\relax PMLR, 2020, pp. 1597--1607.

\bibitem{liu2022contrastive}
A.~Liu, C.~Zhao, Z.~Yu, J.~Wan, A.~Su, X.~Liu, Z.~Tan, S.~Escalera, J.~Xing,
  Y.~Liang \emph{et~al.}, ``Contrastive context-aware learning for 3d
  high-fidelity mask face presentation attack detection,'' \emph{TIFS},
  vol.~17, pp. 2497--2507, 2022.

\bibitem{patchnce}
T.~Park, A.~A. Efros, R.~Zhang, and J.-Y. Zhu, ``Contrastive learning for
  unpaired image-to-image translation,'' in \emph{ECCV}.\hskip 1em plus 0.5em
  minus 0.4em\relax Springer, 2020, pp. 319--345.

\bibitem{dclgan}
J.~Han, M.~Shoeiby, L.~Petersson, and M.~A. Armin, ``Dual contrastive learning
  for unsupervised image-to-image translation,'' in \emph{CVPR}, 2021, pp.
  746--755.

\bibitem{contrastive4}
H.~Wu, Y.~Qu, S.~Lin, J.~Zhou, R.~Qiao, Z.~Zhang, Y.~Xie, and L.~Ma,
  ``Contrastive learning for compact single image dehazing,'' in \emph{CVPR},
  2021, pp. 10\,551--10\,560.

\bibitem{contrastive5}
G.~Wu, J.~Jiang, X.~Liu, and J.~Ma, ``A practical contrastive learning
  framework for single image super-resolution,'' \emph{arXiv preprint
  arXiv:2111.13924}, 2021.

\bibitem{resUnet}
A.~Hertz, S.~Fogel, R.~Hanocka, R.~Giryes, and D.~Cohen-Or, ``Blind visual
  motif removal from a single image,'' in \emph{CVPR}, 2019, pp. 6858--6867.

\bibitem{adain}
X.~Huang and S.~Belongie, ``Arbitrary style transfer in real-time with adaptive
  instance normalization,'' in \emph{ICCV}, 2017, pp. 1501--1510.

\bibitem{contrastive3}
A.~Dosovitskiy, J.~T. Springenberg, M.~Riedmiller, and T.~Brox,
  ``Discriminative unsupervised feature learning with convolutional neural
  networks,'' \emph{NIPS}, vol.~27, 2014.

\bibitem{contrastive6}
O.~Henaff, ``Data-efficient image recognition with contrastive predictive
  coding,'' in \emph{ICML}.\hskip 1em plus 0.5em minus 0.4em\relax PMLR, 2020,
  pp. 4182--4192.

\bibitem{cun2020split}
X.~Cun and C.-M. Pun, ``Split then refine: Stacked attention-guided resunets
  for blind single image visible watermark removal,'' \emph{AAAI}, 2021.

\bibitem{pytorch}
A.~Paszke, S.~Gross, S.~Chintala, G.~Chanan, E.~Yang, Z.~DeVito, Z.~Lin,
  A.~Desmaison, L.~Antiga, and A.~Lerer, ``Automatic differentiation in
  pytorch,'' 2017.

\bibitem{adamw}
I.~Loshchilov and F.~Hutter, ``Fixing weight decay regularization in adam,''
  2018.

\bibitem{jiang2021ssh}
Y.~Jiang, H.~Zhang, J.~Zhang, Y.~Wang, Z.~Lin, K.~Sunkavalli, S.~Chen,
  S.~Amirghodsi, S.~Kong, and Z.~Wang, ``Ssh: A self-supervised framework for
  image harmonization,'' in \emph{ICCV}, 2021, pp. 4832--4841.

\bibitem{color_trans1}
U.~Fecker, M.~Barkowsky, and A.~Kaup, ``Histogram-based prefiltering for
  luminance and chrominance compensation of multiview video,'' \emph{TCSVT},
  vol.~18, no.~9, pp. 1258--1267, 2008.

\bibitem{color_trans2}
E.~Reinhard, M.~Adhikhmin, B.~Gooch, and P.~Shirley, ``Color transfer between
  images,'' \emph{IEEE Computer graphics and applications}, vol.~21, no.~5, pp.
  34--41, 2001.

\bibitem{s2crnet}
J.~Liang and C.-M. Pun, ``Spatial-separated curve rendering network for
  efficient and high-resolution image harmonization,'' \emph{arXiv preprint
  arXiv:2109.05750}, 2021.

\bibitem{zhou2018learning}
P.~Zhou, X.~Han, V.~I. Morariu, and L.~S. Davis, ``Learning rich features for
  image manipulation detection,'' in \emph{CVPR}, 2018, pp. 1053--1061.

\end{thebibliography}
\end{document}


\maketitle

\appendix


\section{Comparison with other contrastive learning loss functions}
Contrastive learning loss function~(Eq.\ref{eq:moco}) in \cite{moco} and \cite{simclr} is mainly designed to compute the similarity of the positive and negative samples in unsupervised training. It does not include positive samples in denominator while our proposed method do as we hope the harmonization model would rather choose to generate background styles than foreground styles. Our goal is to make the generated features more like background styles. Practical explanations of Eq.\ref{eq:ours} is that the generated features should be more closely associated with the background style features. The smaller $\mathcal{L}_{C L}$ is, the more closed it will become. From the model perspective, it turns our to be a problem that is to select positive samples from the total set of the positive and negative samples but not only from the negative samples, which are the biggest difference compared to \cite{moco} and \cite{simclr}.
\begin{equation}
\mathcal{L}_{C L}\left(v^{fg}, v^{bg}, \mathcal{P}_{i}, \{\mathcal{N}_i\} \right)=-\log \frac{\exp \left(v^{bg} \cdot \mathcal{P}_{i} / \tau\right)}{\exp \left(v^{bg} \cdot \mathcal{P}_{i} / \tau\right)+\sum_{n=1}^{K} \exp \left(v^{fg} \cdot \mathcal{N}_{n} / \tau\right)},
\label{eq:ours}
\end{equation}

\begin{equation}
\mathcal{L}_{q}=-\log \frac{\exp \left(q \cdot k_{+} / \tau\right)}{\sum_{i=0}^{K} \exp \left(q \cdot k_{i} / \tau\right)}
\label{eq:moco}
\end{equation}

\section{Analyses of The Quantities of Sampling Patches}
Table~\ref{table:ablation_k} compares the performance with different patch sampling numbers~\emph{K} during the sample generation~(Section 3.3.1). Although increasing sampling numbers theoretically improves contrast complexity between foreground and background style features, the ablation results reveal that capturing more than 256 patches degrades harmonization performance with a modest drop at PSNR and SSIM metrics.
We conjecture that as the number of negative samples grows, more complex style representations emerge, and our method does not employ a well-designed feature extractor to obtain unified style representations following~\cite{patchnce} but only uses a shared-~\emph{MLP} to obtain embedded positive and negative vectors~(illustrated in Section 3.3.1). Moreover, when cropping and embedding more patches, the training resources also increase, such as computational cost and processing time. Overall, we set $K=256$ for a trade-off between the harmonization performance and the training complexity.

\begin{table}[tbh!]
\centering
\caption{Ablation studies of patch number \emph{K}.}
\begin{tabular}{@{}cccc@{}}
\toprule
Number of \emph{K}    & MSE↓          & PSNR↑         & SSIM↑          \\ \midrule
128  & 29.06         & 37.00            & 98.92          \\
256  & \textbf{25.90} & \textbf{37.50} & \textbf{99.04} \\
512  & 30.87         & 36.79         & 98.88          \\
1024 & 30.35         & 36.69         & 98.83          \\ \bottomrule
\end{tabular}
\label{table:ablation_k}
\end{table}

\begin{table}[tbh!]
\centering
\caption{Harmonization performance on high-resolution performance. All experiments are evaluate at PSNR metric in different image resolutions(256, 512, 1024). The best are marked as boldface.}
\begin{tabular}{@{}ccccccc@{}}
\toprule
Method    & S2AM  & DoveNet & BargainNet & RainNet & IIH   & Ours  \\ \midrule
256×256   & 35.34 & 34.34   & 35.34      & 36.20   & 35.20 & \textbf{38.33}  \\
512×512   & 32.55 & 34.72   & 35.78      & 36.56   & 36.62 & \textbf{38.97}  \\
1024×1024 & 30.02 & 34.43   & 34.64      & 36.81   & 36.45 & \textbf{37.55} \\ \bottomrule
\end{tabular}
\end{table}

\begin{figure*}[tbh!]
    \centering
    \includegraphics[width=\textwidth]{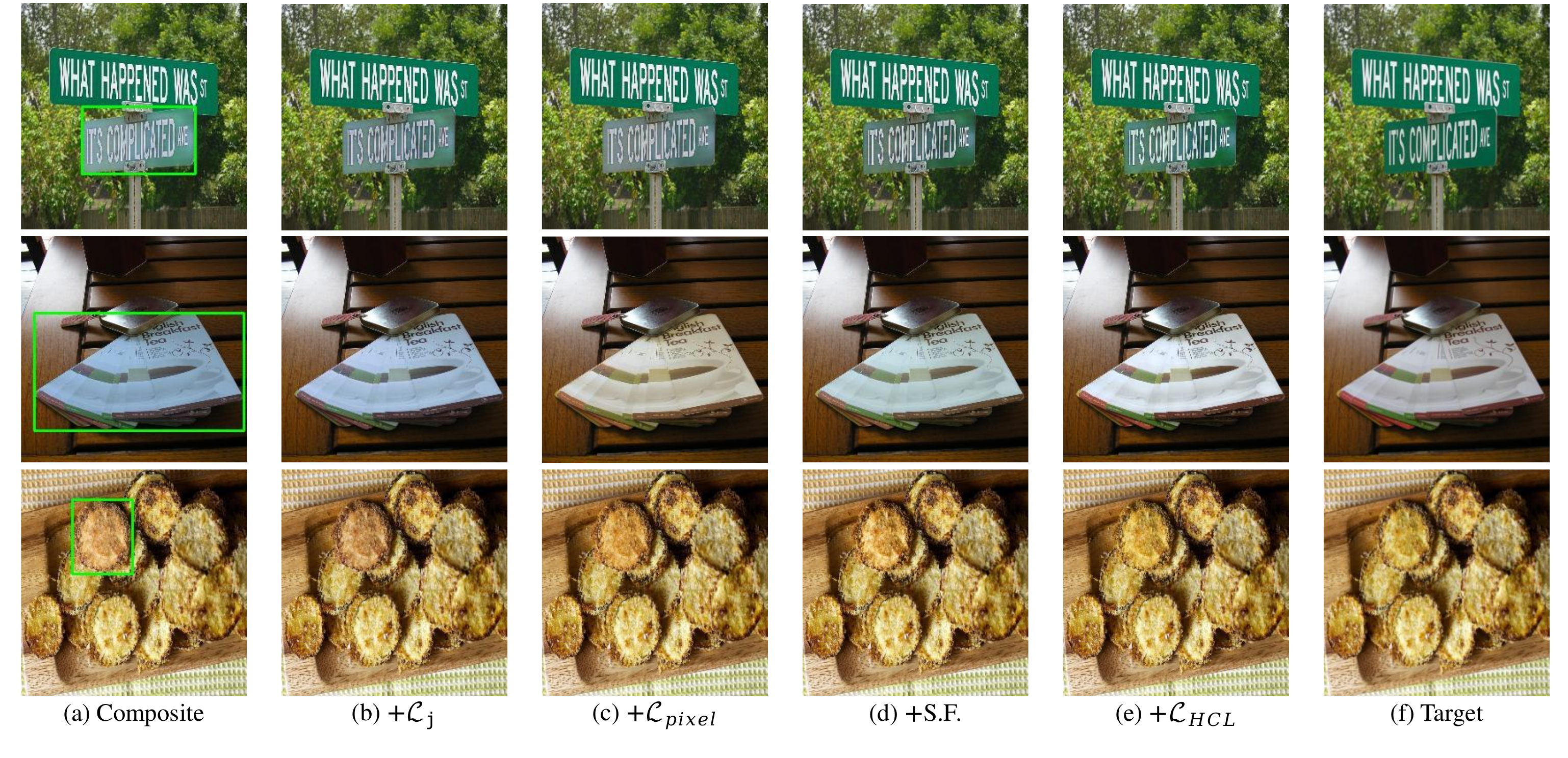}
    \caption{More visualized examples in ablation studies. S.F. denotes our proposed style fusion strategy.}
    \label{fig:abl}
\end{figure*}

\bibliographystyle{abbrv}
\bibliography{ref.bib}